\documentclass[journal]{IEEEtran}
\ifCLASSINFOpdf
\else
\fi
\usepackage{palatino,epsfig,latexsym}
\usepackage{booktabs} 
\usepackage{subfigure,cite}
\usepackage{enumitem}
\usepackage{graphicx}
\usepackage{amsthm,amsmath}
\usepackage{amssymb}
\usepackage{algpseudocode}
\usepackage{algorithm}
\usepackage{color}
\usepackage{url}
\usepackage{makecell,multirow}

\theoremstyle{definition}

\begin{document}

\title{Learning to Approximate: Auto Direction Vector Set Generation for Hypervolume Contribution Approximation}  

\author{Ke~Shang,
	~Tianye~Shu,
        ~Hisao~Ishibuchi,~\IEEEmembership{Fellow,~IEEE}
\thanks{This work was supported by National Natural Science Foundation of China (Grant No. 62002152, 61876075), Guangdong Provincial Key Laboratory (Grant No. 2020B121201001), the Program for Guangdong Introducing Innovative and Enterpreneurial Teams (Grant No. 2017ZT07X386), The Stable Support Plan Program of Shenzhen Natural Science Fund (Grant No. 20200925174447003), Shenzhen Science and Technology Program (Grant No. KQTD2016112514355531). \textit{(Corresponding Author: Hisao Ishibuchi.)}}
\thanks{K. Shang, T. Shu, H. Ishibuchi are with Guangdong Provincial Key Laboratory of Brain-inspired Intelligent Computation, Department of Computer Science and Engineering, Southern University of Science and Technology, Shenzhen 518055, China (e-mail: kshang@foxmail.com; \{12132356@mail., hisao@\}sustech.edu.cn).}
}

\maketitle

\begin{abstract}
Hypervolume contribution is an important concept in evolutionary multi-objective optimization (EMO). It involves in hypervolume-based EMO algorithms and  hypervolume subset selection algorithms. Its main drawback is that it is computationally expensive in high-dimensional spaces, which limits its applicability to many-objective optimization. Recently, an R2 indicator variant (i.e., $R_2^{\text{HVC}}$ indicator) is proposed to approximate the hypervolume contribution. The $R_2^{\text{HVC}}$ indicator uses line segments along a number of direction vectors for hypervolume contribution approximation. It has been shown that different direction vector sets lead to different approximation quality. In this paper, we propose \textit{Learning to Approximate (LtA)}, a direction vector set generation method for the $R_2^{\text{HVC}}$ indicator. The direction vector set is automatically learned from training data. The learned direction vector set can then be used in the $R_2^{\text{HVC}}$ indicator to improve its approximation quality. The usefulness of the proposed LtA method is examined by comparing it with other commonly-used direction vector set generation methods for the $R_2^{\text{HVC}}$ indicator. Experimental results suggest the superiority of LtA over the other methods for generating high quality direction vector sets.
\end{abstract}

\IEEEpeerreviewmaketitle

\section{Introduction}
\IEEEPARstart{I}{n} evolutionary multi-objective optimization (EMO), there are many performance indicators such as generational distance (GD) \cite{van1999multiobjective}, inverted generational distance (IGD) \cite{coello2004study}, hypervolume \cite{zitzler2003performance,shang2020survey}, and R2 \cite{hansen1998evaluating}. Among these performance indicators, the hypervolume indicator is one of the most popular ones. It is able to evaluate the convergence and the diversity of a solution set simultaneously. Thus, it is widely used for performance evaluation of EMO algorithms. On the other hand, it can also be used for EMO algorithm design. Representative hypervolume-based EMO algorithms include SMS-EMOA \cite{Emmerich2005An,beume2007sms}, FV-MOEA \cite{jiang2014simple}, HypE \cite{bader2011hype}, and R2HCA-EMOA \cite{shang2020new}. 

In hypervolume-based EMO algorithms, a multi-objective optimization problem is transformed into a single-objective optimization problem where the single objective is to maximize the hypervolume of the population. In a standard $(\mu+1)$ hypervolume-based EMO algorithm (e.g., SMS-EMOA), in each generation, one offspring is generated and added into the population, then one individual with the least hypervolume contribution is removed from the population. In this manner, the hypervolume of the population is monotonically non-decreasing as the number of generations increases. 

The hypervolume contribution is the key concept in hypervolume-based EMO algorithms. It describes the change of the hypervolume when an individual is added to or removed from the population. Its main drawback is that it is computationally expensive in high-dimensional spaces \cite{bringmann2012approximating}, which limits its applicability to many-objective optimization. In order to overcome this drawback, some hypervolume contribution approximation methods have been proposed \cite{bader2010faster,bringmann2012approximating,shang2018r2,deng2019approximating}. Two representative methods are the point-based method and the line-based method. The point-based method is also known as Monte Carlo method \cite{bader2010faster}. In this method, a large number of points are sampled in the objective space and the hypervolume contribution is approximated based on the percentage of the points lying inside of the hypervolume contribution region. The line-based method is also known as an R2 indicator variant (i.e., $R_2^{\text{HVC}}$ indicator) \cite{shang2018r2}. In this method, a set of line segments with different directions is used to approximate the hypervolume contribution. It has been shown that the line-based method performs better than the point-based method in terms of both runtime and approximation quality \cite{shang2018r2}.  

One important component in the line-based method is the direction vector set, which defines the directions of the line segments. In our previous study \cite{nan2020good}, we investigated the effect of different direction vector set generation methods on the approximation quality of the $R_2^{\text{HVC}}$ indicator. Our results showed that different direction vector sets lead to different approximation quality. A uniform direction vector set has a worse performance than a non-uniform direction vector set, which is counterintuitive. This motivates us to study what is the best direction vector set for the $R_2^{\text{HVC}}$ indicator. 

In this paper, we investigate how to find the best direction vector set for the $R_2^{\text{HVC}}$ indicator. Our proposed method, \textit{Learning to Approximate (LtA)}, is able to automatically train a direction vector set from training data. The trained direction vector set can then be used in the $R_2^{\text{HVC}}$ indicator to improve its approximation quality. The main contributions of this paper are summarized as follows.
\begin{enumerate}
\item We formulate the task of finding the best direction vector set as an optimization problem. The objective function is to maximize the Pearson correlation coefficient between the true and the approximated hypervolume contribution values for training data. The training data is a number of non-dominated solution sets.
\item We propose the LtA method to solve the defined optimization problem. The LtA method is a population-based algorithm where each individual is a direction vector. The population evolves in a steady-state manner which guarantees that the objective is monotonically non-decreasing as the number of generations increases.
\item We verify the effectiveness of the LtA method by comparing it with other commonly-used direction vector set generation methods based on three different experiments, which allow us to evaluate the performance of the proposed method from different perspectives. 
\end{enumerate}

The rest of the paper is organized as follows. Section II presents the preliminaries of the study. Section III introduces the proposed method, LtA, to automatically generate a direction vector set. Section IV conducts the experimental studies. Section V gives further investigations, and Section VI concludes the paper. 

\section{Preliminaries}
\subsection{Basic Definitions}
In evolutionary multi-objective optimization (EMO), the hypervolume indicator is defined as follows. Given a solution set $S$ in the objective space, the hypervolume of $S$ is defined as
\begin{equation}
HV(S,\mathbf{r}) = \mathcal{L}\left(\bigcup_{\mathbf{s}\in S}\left\{\mathbf{s}'|\mathbf{s}\prec \mathbf{s}'\prec \mathbf{r}\right\}\right),
\end{equation}
where $\mathcal{L}(.)$ is the Lebesgue measure of a set, $\mathbf{r}$ is the reference point which is dominated by all solutions in $S$, and $\mathbf{s}\prec \mathbf{s}'$ denotes that $\mathbf{s}$ Pareto dominates $\mathbf{s}'$ (i.e., $s_i\leq s'_i$ for all $i=1,...,m$ and $s_j<s'_j$ for at least one $j=1,...,m$ in the minimization case, where $m$ is the number of objectives).

\begin{figure}[!htbp]
\centering                                           
\subfigure[Hypervolume]{ 
\includegraphics[scale=0.33]{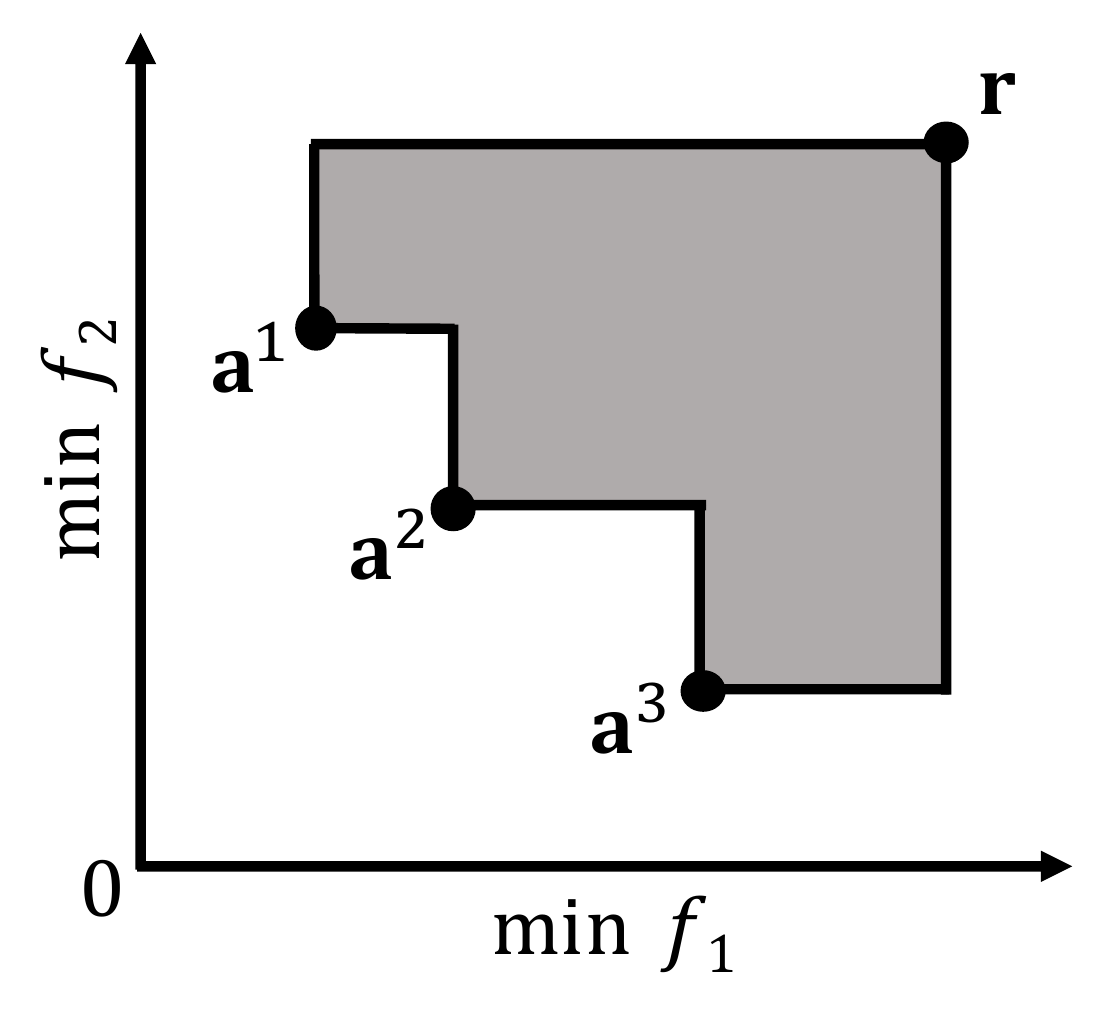}   }            
\subfigure[Hypervolume Contribution]{ 
\includegraphics[scale=0.33]{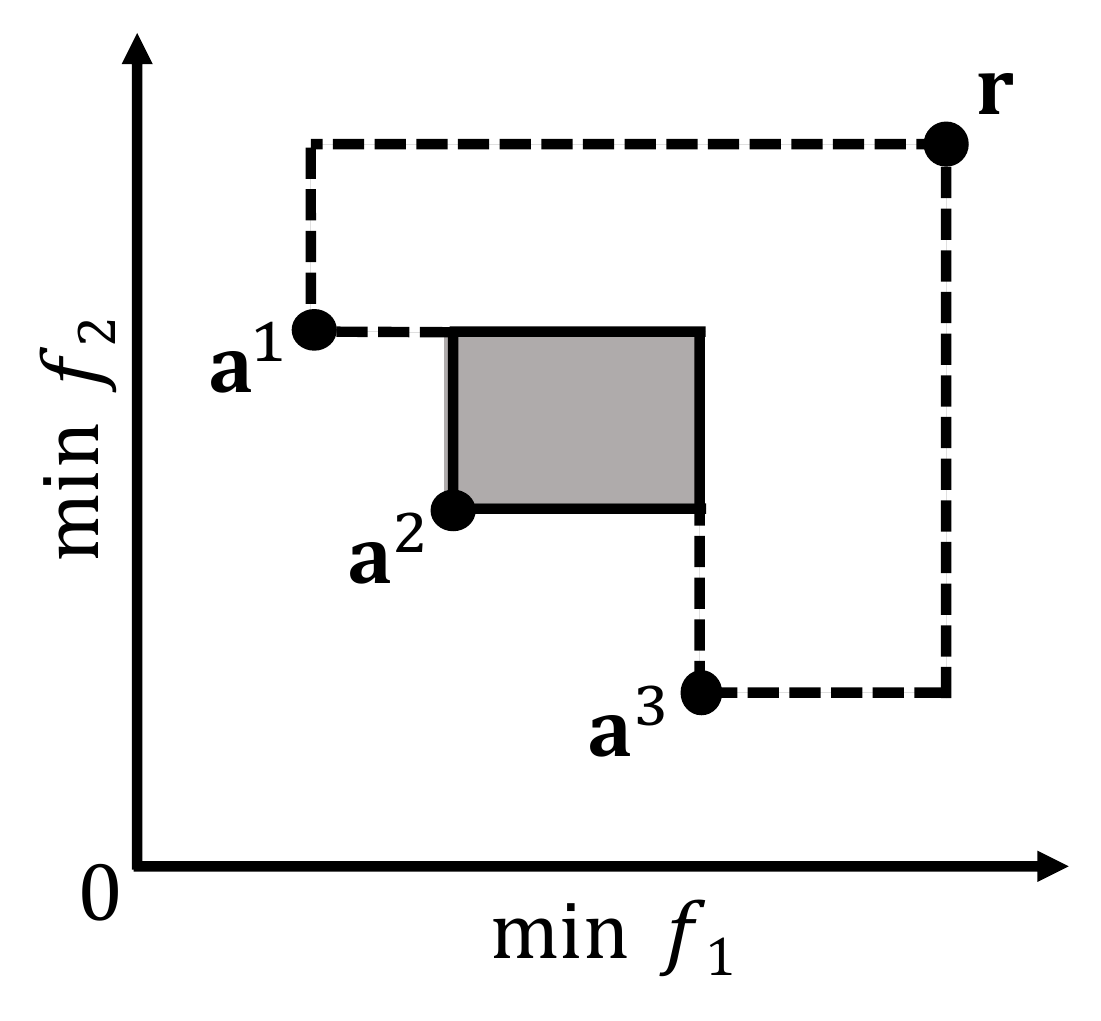} }
\caption{Illustrations of the hypervolume and the hypervolume contribution. The shaded area in (a) is the hypervolume of the solution set $S= \{\mathbf{a}^1,\mathbf{a}^2,\mathbf{a}^3\}$, and the shaded area in (b) is the hypervolume contribution of solution $\mathbf{a}^2$ to the solution set $S = \{\mathbf{a}^1,\mathbf{a}^3\}$. } 
\label{hv}                                                        
\end{figure}

Fig. \ref{hv} (a) illustrates the hypervolume of a solution set $S = \{\mathbf{a}^1,\mathbf{a}^2,\mathbf{a}^3\}$ in a two-dimensional objective space, where each objective is to be minimized. 

The hypervolume contribution is defined based on the hypervolume indicator. It describes the change of the hypevolume when a solution is added to or removed from the solution set. Formally, given a solution $\mathbf{s}\in S$, the hypervolume contribution of $\mathbf{s}$ to $S$ is defined as
\begin{equation}
HVC(\mathbf{s},S,\mathbf{r}) = HV(S,\mathbf{r}) - HV(S\setminus \{\mathbf{s}\},\mathbf{r}),
\end{equation}
If $\mathbf{s}\notin S$, the hypervolume contribution of $\mathbf{s}$ to $S$ is defined as
\begin{equation}
HVC(\mathbf{s},S,\mathbf{r}) = HV(S\cup \{\mathbf{s}\},\mathbf{r}) - HV(S,\mathbf{r}).
\end{equation}

Fig. \ref{hv} (b) illustrates the hypervolume contribution of solution $\mathbf{a}^2$ to the solution set $S = \{\mathbf{a}^1,\mathbf{a}^3\}$. 

\subsection{Hypervolume Contribution Approximation}
In \cite{shang2018r2}, an R2 indicator variant (i.e., $R_2^{\text{HVC}}$ indicator)  is proposed to approximate the hypervolume contribution of a solution $\mathbf{s}$ to a  solution set $S$ in an $m$-dimensional objective space:
\begin{equation}
\begin{aligned}
\label{newmethod}
&R^{\text{HVC}}_2(\mathbf{s},S,\mathbf{r})\\
&=\frac{1}{|\Lambda|}\sum_{{\boldsymbol{\lambda}}\in \Lambda}\min\left\{\min_{\mathbf{s}'\in S}\left\{g^{\text{*2tch}}(\mathbf{s}'|{\boldsymbol{\lambda}},\mathbf{s})\right\},g^{\text{mtch}}(\mathbf{r}|{\boldsymbol{\lambda}},\mathbf{s})\right\}^m,
\end{aligned}
\end{equation}
where $S$ is a non-dominated solution set and $\mathbf{s}\notin S$, $\Lambda$ is a direction vector set, each direction vector ${\boldsymbol{\lambda}}=(\lambda_1,\lambda_2,...,\lambda_m)\in\Lambda$ satisfies $\left \| {\boldsymbol{\lambda}} \right \|_2 = 1$ and $\lambda_i\geq 0$ for $i = 1,...,m$, and $\mathbf{r}$ is the reference point. 

The $g^{\text{*2tch}}(\mathbf{s}'|{\boldsymbol{\lambda}},\mathbf{s})$ function in \eqref{newmethod} is defined for minimization problems as 
\begin{equation}
g^{\text{*2tch}}(\mathbf{s}'|{\boldsymbol{\lambda}},\mathbf{s})=\max_{j\in\{1,...,m\}}\left\{\frac{s'_j-s_j}{\lambda_j}\right\}.
\end{equation}
For maximization problems, it is defined as
\begin{equation}
\label{g2tchnew}
g^{\text{*2tch}}(\mathbf{s}'|{\boldsymbol{\lambda}},\mathbf{s})=\max_{j\in\{1,...,m\}}\left\{\frac{s_j-s'_j}{\lambda_j}\right\}.
\end{equation}

The $g^{\text{mtch}}$ function in \eqref{newmethod} is defined for both minimization and maximization problems as
\begin{equation}
\label{gmtchnew}
g^{\text{mtch}}(\mathbf{r}|{\boldsymbol{\lambda}},\mathbf{s})=\min_{j\in\{1,...,m\}}\left\{\frac{|s_j-r_j|}{\lambda_j}\right\}.
\end{equation}

\begin{figure}
\centering                                           
\includegraphics[scale=0.33]{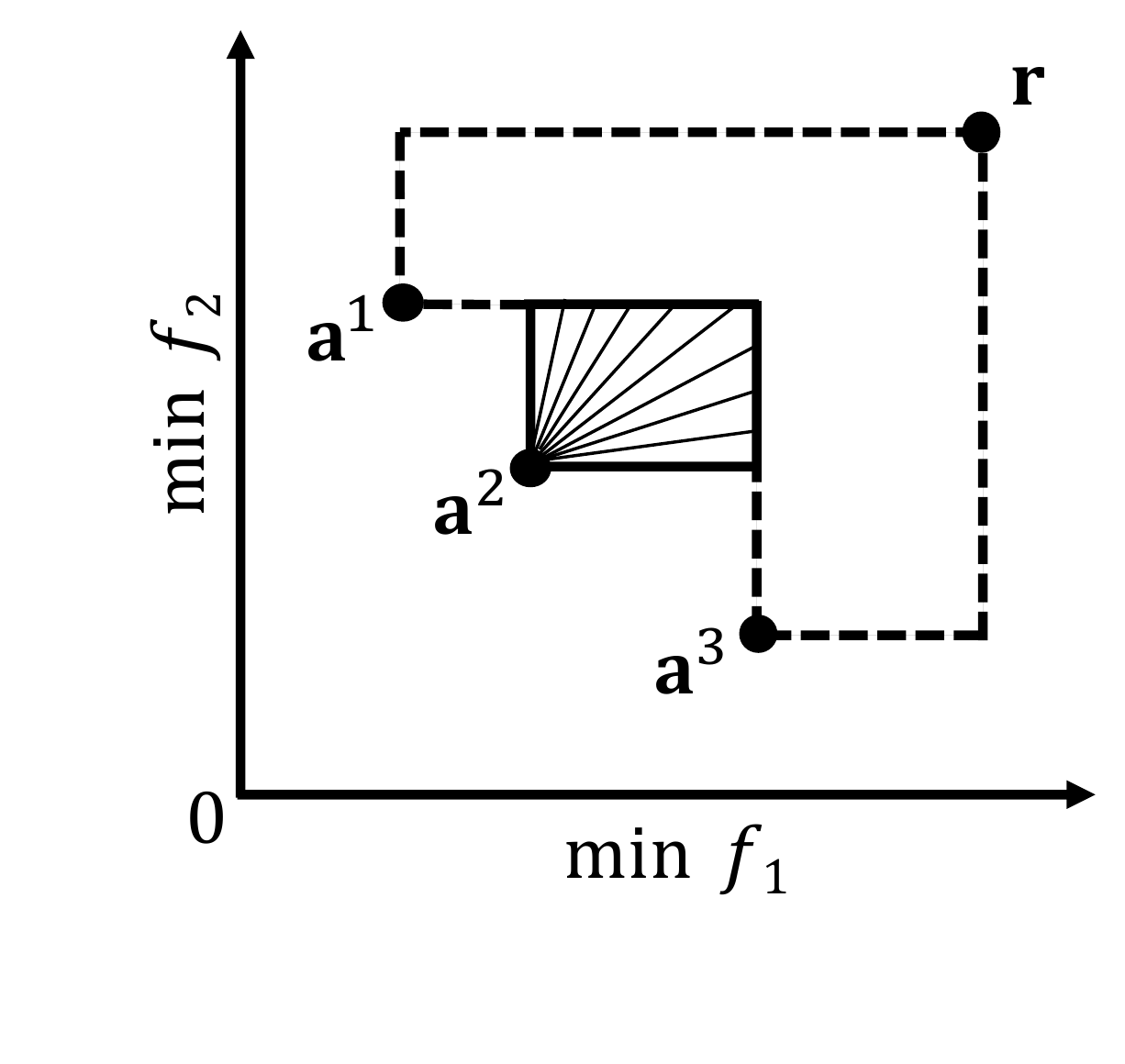}               
\caption{An illustration of the geometric meaning of the $R_2^{\text{HVC}}$ indicator for the hypervolume contribution approximation.} 
\label{fig:r2}
\end{figure}

The geometric meaning of the $R_2^{\text{HVC}}$ indicator is illustrated in Fig. \ref{fig:r2}. Given a non-dominated solution set $S = \{\mathbf{a}^1,\mathbf{a}^3\}$ and a set of line segments with different directions in the hypervolume contribution region of $\mathbf{a}^2$, the $R_2^{\text{HVC}}$ indicator in \eqref{newmethod} is calculated for $\mathbf{a}^2$ as  $R^{\text{HVC}}_2(\mathbf{a}^2,S,\mathbf{r})=\frac{1}{|\Lambda|}\sum_{i=1}^{|\Lambda|} l_i^m$ where $l_i$ is the length of the $i$th line segment. For more detailed explanations of the $R_2^{\text{HVC}}$ indicator, please refer to \cite{shang2018r2}.

\subsection{Direction Vector Set Generation Methods}
\label{generationmethods}
In the $R_2^{\text{HVC}}$ indicator in \eqref{newmethod}, a direction vector set $\Lambda$ is used to specify the directions of the line segments for the hypervolume contribution approximation. Intuitively, one may think that a uniformly distributed direction vector set is a good choice. However, this intuition is not correct. In our previous study in \cite{nan2020good}, we investigated the effects of five different direction vector set generation methods on the performance of the $R_2^{\text{HVC}}$ indicator. Our experimental results showed that a uniform direction vector set has a worse performance than a non-uniform direction vector set.

The five generation methods considered in \cite{nan2020good} are briefly explained as follows.
\subsubsection{DAS}
The Das and Dennis's (DAS) method \cite{das1998normal} is widely used in decomposition-based EMO algorithms (e.g., MOEA/D \cite{zhang2007moea}, NSGA-III \cite{deb2013evolutionary}) to generate weight vectors. DAS generates all weight vectors $\mathbf{w}=(w_1,w_2,...,w_m)$ satisfying the following relations:
\begin{equation}
\begin{aligned}
&\sum_{i=1}^{m}w_i=1 \text{ and } w_i\in\left\{0,\frac{1}{H},\frac{2}{H},...,1\right\} \text{ for }i=1,2,...,m,
\end{aligned}
\end{equation}
where $m$ is the number of objectives and $H$ is a positive integer. The total number of generated weight vectors is $\binom{H+m-1}{m-1}$.

Based on the weight vector $\mathbf{w}$, we obtain the direction vector $\boldsymbol{\lambda}=\mathbf{w}/\left \| {\mathbf{w}} \right \|_2$.

\subsubsection{UNV}
The unit normal vector (UNV) method \cite{deng2019approximating} is a sampling method which samples points uniformly on the unit hypersphere. The direction vectors are generated by uniformly sampling points on the unit hypersphere as follows. First we randomly sample points $\mathbf{x}$ (i.e., $m$-dimensional vectors) according to the $m$-dimensional normal distribution $\mathcal{N}_m(0,I_m)$, then the corresponding direction vectors are obtained by $\boldsymbol{\lambda} = |\mathbf{x}|/\left \| \mathbf{x} \right \|_2$. 

\begin{figure}[!htbp]
\centering                                           
\subfigure[DAS]{ 
\includegraphics[scale=0.3]{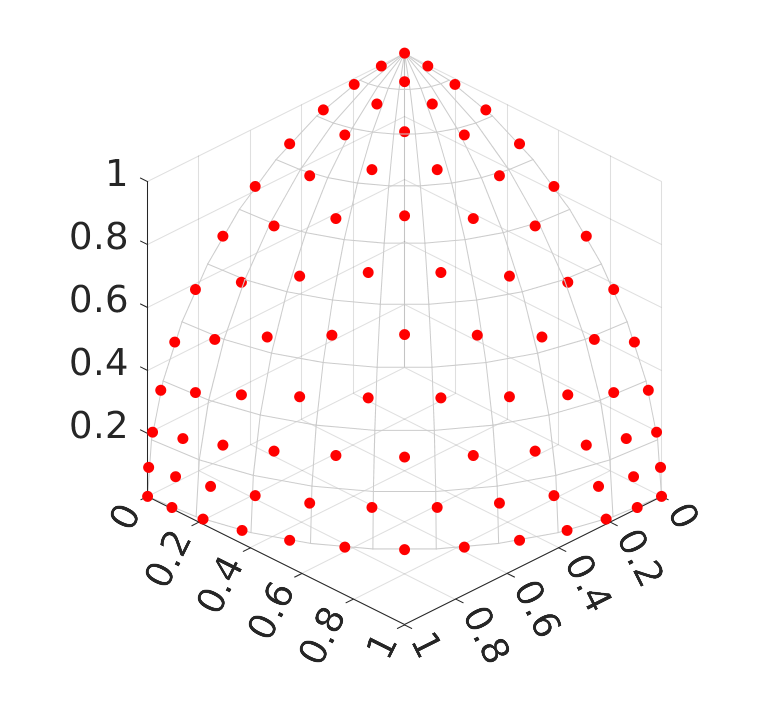}   }            
\subfigure[UNV]{ 
\includegraphics[scale=0.3]{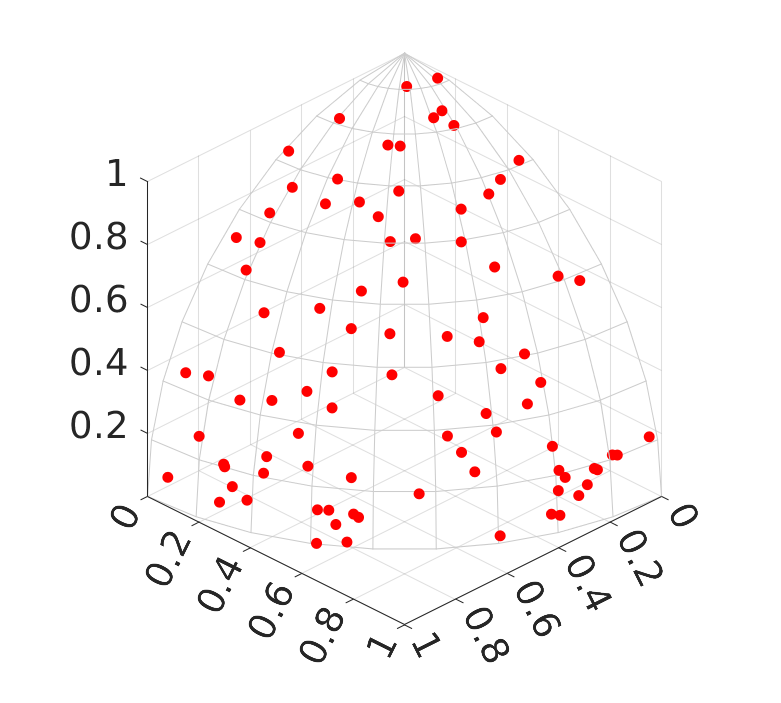} }
\subfigure[JAS]{ 
\includegraphics[scale=0.3]{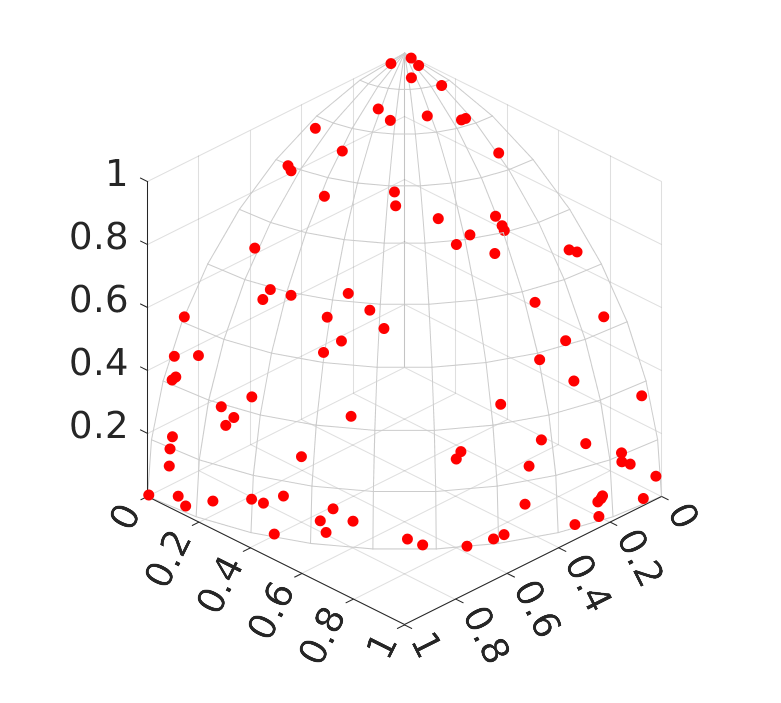} }
\subfigure[MSS-D]{ 
\includegraphics[scale=0.3]{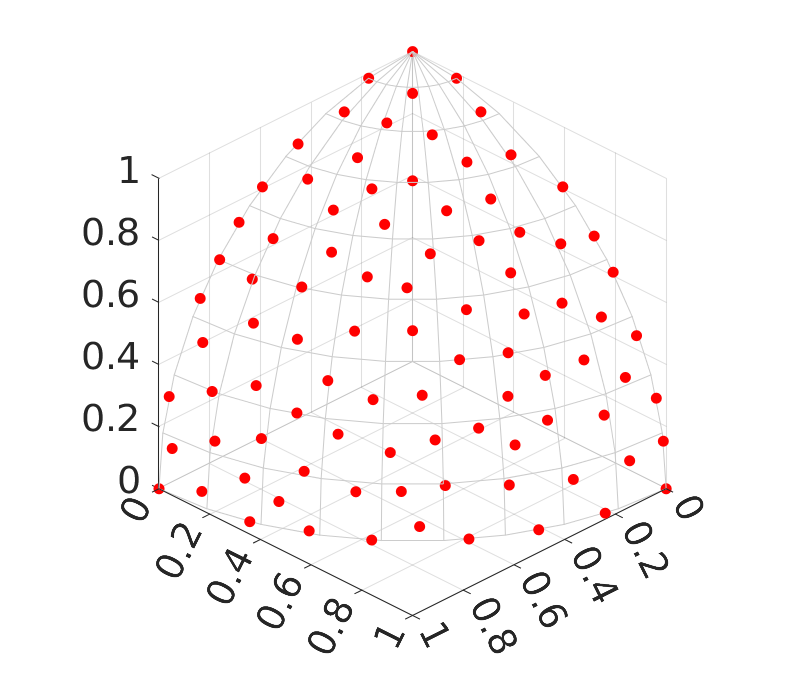} }
\subfigure[MSS-U]{ 
\includegraphics[scale=0.3]{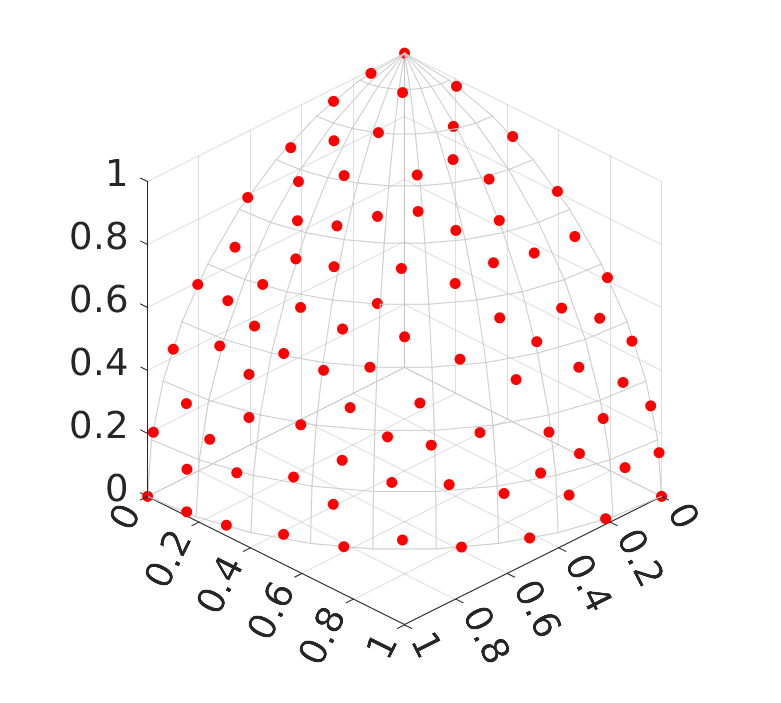} }
\subfigure[Kmeans-U]{ 
\includegraphics[scale=0.3]{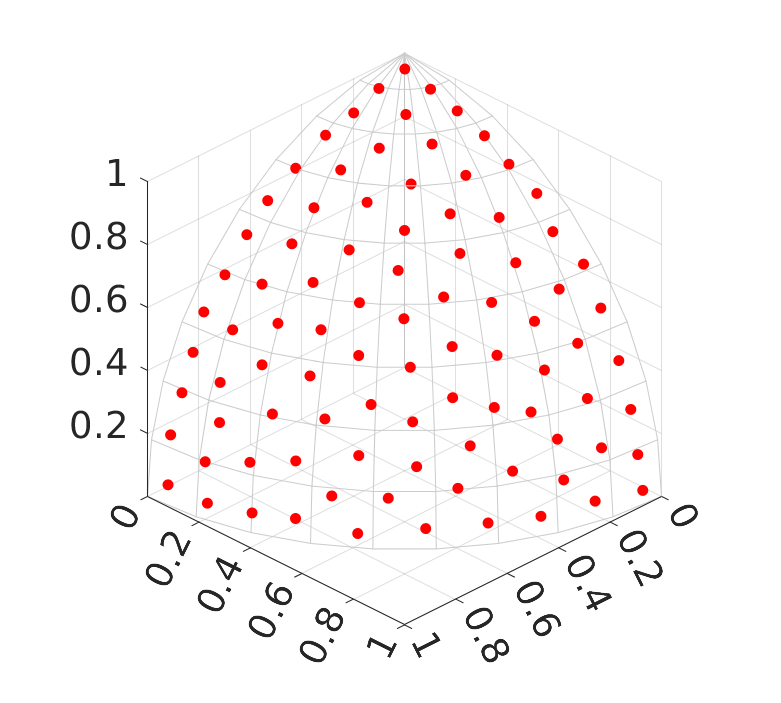} }
\caption{The distributions of the direction vector sets generated by six methods described in Section \ref{generationmethods}.} 
\label{5dvs}                                                        
\end{figure}

\subsubsection{JAS}
The JAS method \cite{jaszkiewicz2002performance} is a probabilistic filling method for generating a set of weight vectors on the simplex. A weight vector $\mathbf{w}$ is generated as follows. First choose a random number $\mu_1\in[0,1]$ and compute $w_1=1-\sqrt[m-1]{\mu_1}$. Then choose another random number $\mu_k\in[0,1]$ and compute $w_k = (1-\sum_{j=1}^{k-1}w_j)(1-\sqrt[m-k]{u_k})$ for $k=2,...,m-1$. Lastly, $w_m$ is computed as $w_m=1-\sum_{j=1}^{m-1}w_j$. This procedure is iterated to create a pre-specified number of weight vectors.

Based on the weight vector $\mathbf{w}$, we obtain the direction vector $\boldsymbol{\lambda}=\mathbf{w}/\left \| {\mathbf{w}} \right \|_2$.

\subsubsection{MSS-D}
The maximally sparse selection (MSS) method  \cite{deb2019generating} starts with a large point set $S$ on the unit hypersphere. First, the direction vector set $\Lambda$ is initialized as the $m$ extreme points $(1,0,...,0)$, $(0,1,...,0)$, ..., $(0,0,...,1)$. Then the point in $S$ with the largest distance to $\Lambda$ is selected and added to $\Lambda$. This procedure is repeated until a pre-specified number of direction vectors are included in $\Lambda$. If $S$ is generated by the DAS method, the MSS method is called MSS-D. 

\subsubsection{MSS-U}
When $S$ is generated by the UNV method, the MSS method is called MSS-U.

In addition to the above five methods, we also consider the following method in this paper.
\subsubsection{Kmeans-U}
This method is described in \cite{ishibuchi2018reference}. At first, a large point set $S$ on the unit hypersphere is generated by the UNV method. Then, the k-means clustering \cite{macqueen1967some} is applied to $S$ to select a pre-specified number of points as the direction vector set. In the k-means clustering, the number of clusters is the same as the number of direction vectors to be selected. The closest point to each cluster center is selected. 

Fig. \ref{5dvs} (a)-(f) show 91 direction vectors generated by each method in the three-dimensional objective space. We can see that the UNV and JAS methods generate non-uniform direction vectors whereas uniform direction vectors are generated by the other methods. However, it was reported in \cite{nan2020good} that the UNV and JAS methods have better approximation quality than DAS, MSS-D and MSS-U. In this paper, all the six generation methods of direction vectors in Fig. \ref{5dvs}  are compared with the proposed method. 

Fig. \ref{5dvs} (f) shows 91 direction vectors generated by Kmeans-U.  As shown in Fig. \ref{5dvs} (f), the direction vectors generated by this method are uniformly distributed. However, different from the DAS, MSS-D, and MSS-U methods, no boundary direction vector is generated by Kmeans-U. In fact, the direction vector set distribution in Fig. \ref{5dvs} (f) is closely related to IGD minimization when $S$ is used as the reference point set \cite{ishibuchi2018reference}.

\section{Learning to Approximate}
\subsection{Mathematical Modeling}
Our task is to find a direction vector set which has high approximation quality for the $R_2^{\text{HVC}}$ indicator. Mathematically, we can formulate this task as an optimization problem as follows.
\begin{equation}
\label{optimization}
\begin{aligned}
\text{Maximize: }&Q(\Lambda),\\
\text{Subject to: }&\Lambda = \{\boldsymbol{\lambda}^i| i=1,...,n\},\  \left \| {\boldsymbol{\lambda}^i} \right \|_2 = 1,\\
&\lambda^i_j\geq 0, \ i=1,...,n, \ j=1,...,m,
\end{aligned}
\end{equation} 
where $Q$ is a function which can evaluate the approximation quality of the direction vector set $\Lambda$, $n$ is the number of direction vectors in $\Lambda$, and $m$ is the number of objectives.

Now we have two questions with respect to the optimization problem in \eqref{optimization}:
\begin{itemize}
\item Q1: How to define the objective function $Q$?
\item Q2: How to solve the optimization problem in \eqref{optimization}? 
\end{itemize}

If the above two questions can be solved, we can generate a good direction vector set for the $R_2^{\text{HVC}}$ indicator. In the next two subsections, we present our solutions with respect to the above two questions, respectively.

\begin{figure}[!htbp]
\centering                                           
\subfigure[$Q=0.2322$]{ 
\includegraphics[scale=0.26]{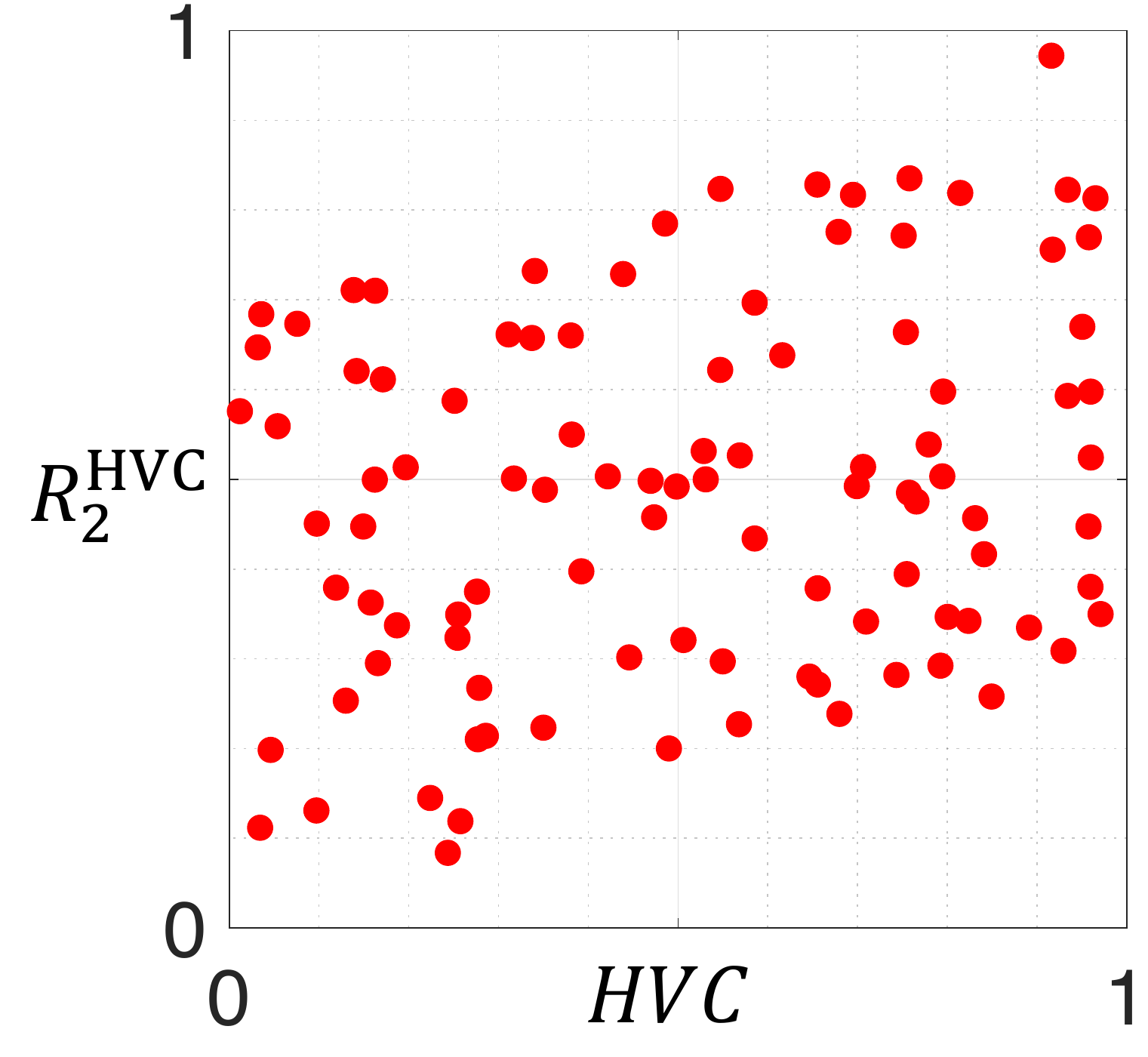}   }            
\subfigure[$Q=0.6603$]{ 
\includegraphics[scale=0.26]{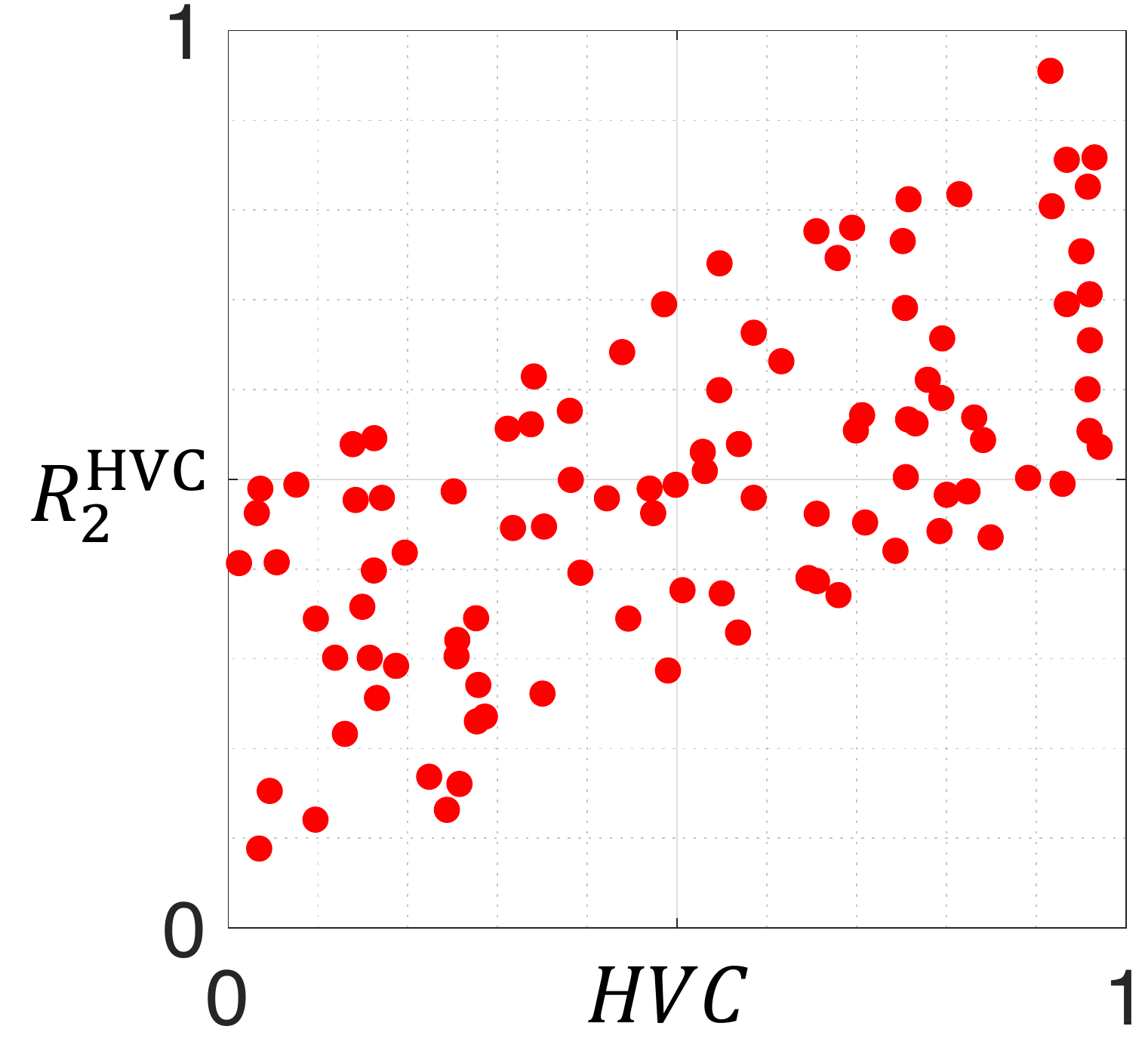} }
\subfigure[$Q=0.9915$]{ 
\includegraphics[scale=0.26]{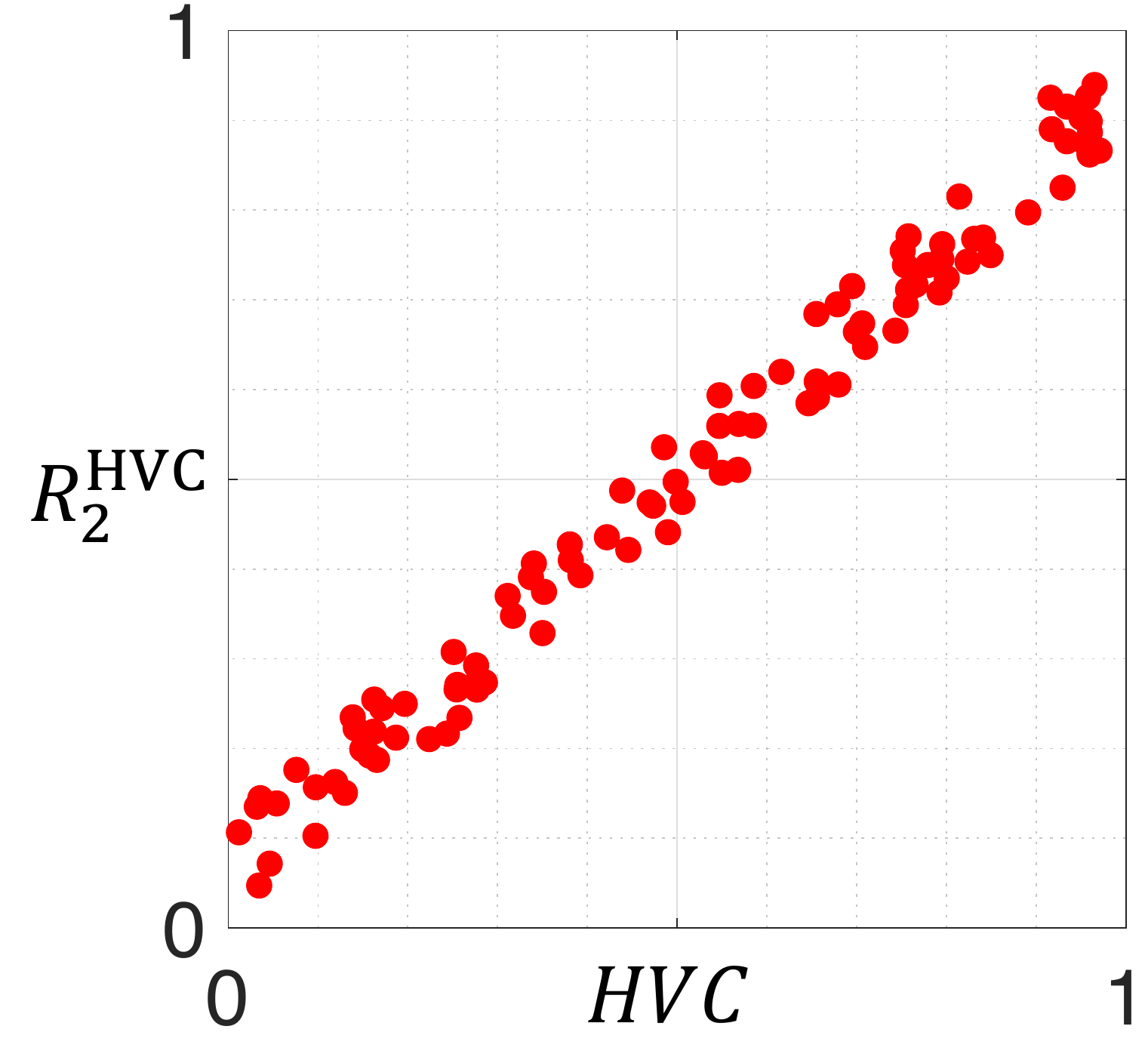} }
\caption{Illustrations of the $Q$ function in \eqref{qfunction} with different data sets.} 
\label{pearson}                                                        
\end{figure}
\subsection{How to Define the Objective Function $Q$?}
The objective function $Q$ is used to evaluate the approximation quality of the direction vector set $\Lambda$. However, we can not evaluate the quality of $\Lambda$ without data (i.e., a given set of solutions with their true and approximated hypervolume contribution values). Thus, we need to prepare data in order to define the objective function $Q$.

Suppose we have a non-dominated solution set $S=\{\mathbf{s}_1,\mathbf{s}_2,...,\mathbf{s}_N\}$, we can calculate the hypervolume contribution and the $R^{\text{HVC}}_2$ value of each solution in $S$, and obtain a set of approximation results as follows in an $N\times2$ matrix form:
\begin{equation}
D(S)=\begin{bmatrix}
HVC(\mathbf{s}_1,S,\mathbf{r}) & R^{\text{HVC}}_2(\mathbf{s}_1,S,\mathbf{r})\\ 
HVC(\mathbf{s}_2,S,\mathbf{r}) & R^{\text{HVC}}_2(\mathbf{s}_2,S,\mathbf{r})\\ 
\vdots  &\vdots  \\ 
HVC(\mathbf{s}_N,S,\mathbf{r}) & R^{\text{HVC}}_2(\mathbf{s}_N,S,\mathbf{r})
\end{bmatrix}.
\end{equation}


If the $R_2^{\text{HVC}}$ indicator is a perfect approximator, we can expect that the relation between $HVC$ and $R^{\text{HVC}}_2$ is strictly linear. However, due to the approximation error, this relation can hardly hold. Therefore, one intuitive way is to use the degree of linearity between $HVC$ and $R^{\text{HVC}}_2$ as the objective function $Q$. The Pearson correlation coefficient \cite{benesty2009pearson} can be used for this purpose. Given the data set $D(S)$, we define the objective function $Q$ as follows
\begin{equation}
\begin{aligned}
\label{qfunction}
Q(\Lambda) = \frac{\sum_{i=1}^N(D_{i,1}-\overline{D}_{:,1})(D_{i,2}-\overline{D}_{:,2})}{\sqrt{\sum_{i=1}^N(D_{i,1}-\overline{D}_{:,1})^2}\sqrt{\sum_{i=1}^N(D_{i,2}-\overline{D}_{:,2})^2}},
\end{aligned}
\end{equation}
where $\overline{D}_{:,1}$ and $\overline{D}_{:,2}$ represent the average values of the first and second columns of $D(S)$, respectively. 

The value of $Q$ lies in $[-1,1]$ where a higher value indicates a better linear relation between $HVC$ and $R^{\text{HVC}}_2$. We can then optimize the direction vector set $\Lambda$ in order to maximize $Q$. Fig. \ref{pearson} illustrates the $Q$ values of different data sets. In Fig. \ref{pearson} (a), there is no clear linear relation between  $HVC$ and $R^{\text{HVC}}_2$, and $Q$ has a low value 0.2322. In Fig. \ref{pearson} (b), the linear relation between $HVC$ and $R^{\text{HVC}}_2$ becomes clearer, and $Q$ is increased to 0.6603. In  Fig. \ref{pearson} (c), we can observe a good linear relation between $HVC$ and $R^{\text{HVC}}_2$, and $Q$ reaches to a high value 0.9915.

\subsection{How to Solve the Optimization Problem in  \eqref{optimization}?}
\label{section:lta}
We can see that \eqref{optimization}  is a set-based optimization problem where a direction vector set is to be optimized. It is not easy to use traditional mathematical programming methods to solve it. A natural idea is to use population-based methods where each individual in the population is a direction vector. Motivated by SMS-EMOA \cite{Emmerich2005An,beume2007sms}, a classical EMO algorithm which aims to maximize the hypervolume of the population, we develop a population-based algorithm for solving \eqref{optimization}. The proposed method, named \textit{Learning to Approximate (LtA)}, is shown in Algorithm \ref{solgeneration}.

\renewcommand{\algorithmicrequire}{\textbf{Input:}} 
\renewcommand{\algorithmicensure}{\textbf{Output:}} 
\begin{algorithm}[h] 
\begin{algorithmic}[1]
\Require 
Solution set $S$, direction vector set size $n$ (i.e., $|\Lambda| = n$), maximum iteration number $maxIteration$, reference point $\mathbf{r}$.
\Ensure 
Direction vector set $\Lambda$.
\State Calculate the hypervolume contribution of each solution in $S$ based on reference point $\mathbf{r}$.
\State Generate $n$ direction vectors using the UNV method to initialize $\Lambda$.
\State $iteration=0$.
\While{$iteration<maxIteration$}
\State Generate a direction vector $\boldsymbol{\lambda}$ using the UNV 
\Statex \ \ \ \ \  method.
\State Update $\Lambda$ as $\Lambda = \Lambda\cup \{\boldsymbol{\lambda}\}$.
\For{each $\boldsymbol{\lambda}' \in \Lambda$}
\State Calculate the $R^{\text{HVC}}_2$ value of each solution in 
\Statex \qquad \ \ \ $S$ based on $\Lambda\setminus \{\boldsymbol{\lambda}'\}$ and reference point $\mathbf{r}$.
\State Obtain data set $D(S)$ and calculate $Q(\Lambda\setminus \{\boldsymbol{\lambda}'\})$ 
\Statex \qquad \ \ \ according to \eqref{qfunction}.
\EndFor
\State $\boldsymbol{\lambda}^* = \arg \max_{\boldsymbol{\lambda}'\in\Lambda} Q(\Lambda\setminus \{\boldsymbol{\lambda}'\})$.
\State $\Lambda = \Lambda\setminus \{\boldsymbol{\lambda}^*\}$.
\State $iteration=iteration+1$.
\EndWhile
\State Return $\Lambda$.
\end{algorithmic} 
\caption{Learning to Approximate (LtA)} 
\label{solgeneration}
\end{algorithm}

In Algorithm \ref{solgeneration}, the hypervolume contribution of each solution in solution set $S$ is calculated (Line 1). Then, the direction vector set $\Lambda$ is initialized using the UNV method (Line 2). After that, the direction vector set $\Lambda$ is optimized in a steady-state manner for a maximum iteration number (Lines 4-14). In each iteration, one direction vector is generated using the UNV method and added into $\Lambda$ (Lines 5-6). Now there are $n+1$ direction vectors in $\Lambda$. We need to remove one direction vector from $\Lambda$. The basic idea is to remove the direction vector so that the rest direction vectors in $\Lambda$ can achieve the best objective function value (i.e., maximum $Q$ value). In order to do this, for each direction vector in $\Lambda$, the corresponding $Q$ value after tentatively removing it is calculated (Lines 7-19). Lastly, the direction vector which leads to the best $Q$ value by its removal is identified and removed from $\Lambda$ (Lines 11-12).

We name Algorithm \ref{solgeneration} as Learning to Approximate (LtA) since $\Lambda$ is learned based on a solution set $S$. From $S$ we can obtain data set $D(S)$, which is used to evaluate $\Lambda$ by calculating $Q(\Lambda)$. It is clear that a single solution set $S$ is not enough to learn a good direction vector set for various multi-objective problems since it is likely that the learned direction vector set is overfitted to $S$ (i.e., the direction vector set has a good $Q$ value for solution set $S$ but not for other solution sets). In order to solve this issue, we improve Algorithm \ref{solgeneration} as follows.
\begin{enumerate}
\item Prepare a number of different solution sets $\{S_1,S_2,...,S_L\}$ instead of a single solution set $S$.
\item For these $L$ solution sets, we can get $L$ data sets $\{D(S_1),D(S_2),...,D(S_L)\}$.
\item For these $L$ data sets, we can calculate $L$ objective function values $\{Q_1(\Lambda),Q_2(\Lambda),...,Q_L(\Lambda)\}$.
\item Calculate $Q(\Lambda)$ as $Q(\Lambda) = \frac{1}{L}\sum_{i=1}^{L}Q_i(\Lambda)$ in Line 9.
\end{enumerate}

Based on the above improvement, we can expect that the learned direction vector set is not overfitted to a specific solution set and has a robust performance across different solution sets. 

\subsection{Discussions}
We have the following discussions with respect to the LtA method.
\subsubsection{How to prepare training solution sets for LtA?}
For the training solution sets $\{S_1,S_2,...,S_L\}$, we need to specify two parameters: $L$ (the number of solution sets) and $N$ (the number of solutions in each solution set). These two parameters determine the overall size of the training solution sets.  In this paper, we set $L=100$ and $N=100$.

In principle, any solution sets can be used for training. In order to make the training more efficient and robust, these $L$ solution sets need to be as diverse as possible. We will describe how to generate diverse solution sets for training in Section \ref{learninglta}.

\subsubsection{Why is Pearson correlation coefficient used as the objective function $Q$?} 
There are many ways to define the approximation quality of the $R_2^{\text{HVC}}$ indicator. We choose Pearson correlation coefficient as the objective function $Q$ since it intuitively defines the linearity between the true and approximated hypervolume contribution values. The other reason is that it is sensitive to the change of the direction vector set. As shown in Algorithm  \ref{solgeneration}, in each iteration, the direction vector set is slightly changed (i.e., only one direction vector is changed and the other direction vectors remain unchanged). If the objective function $Q$ is insensitive to the slight change of the direction vector set, the training will become difficult. Using  Pearson correlation coefficient as the objective function $Q$, the training of the direction vector set is very efficient. 

\subsubsection{What is the time complexity of LtA?}
\label{timecomplexity}
The time complexity of LtA is analyzed as follows. Let us consider a single training solution set $S$ in LtA. The main time cost is the while loop in Algorithm \ref{solgeneration}. In the while loop, the most time-consuming part is the for loop (Line 7-10). In order to efficiently calculate the $R^{\text{HVC}}_2$ value in Line 8, we use the tensor structure introduced in \cite{shang2020new}. The tensor structure stores the information for the calculation of $R^{\text{HVC}}_2$. For the detailed explanation of the tensor structure, please refer to \cite{shang2020new}.

The size of the tensor structure is $N\times (n+1)\times N$ where $N$ is the size of the solution set $S$ and  $n$ is the size of the direction vector set $\Lambda$. The tensor structure is updated in $O(N\times N\times m)$ once a new direction vector is generated (Line 6). Before enter the for loop, the minimum value of each column of the tensor structure is calculated in $O(N\times n\times N)$ for further use. In each iteration of the for loop, the $R^{\text{HVC}}_2$ values can be obtained in $O(n\times N)$ (Line 8). Thus, the for loop is run in $O(n\times n\times N)$. In summary, the time complexity of one iteration of LtA is $O(N^2m+N^2n+n^2N)$.

If we have $L$ training solution sets, the time complexity of one iteration of LtA is $O(L(N^2m+N^2n+n^2N))$.


\section{Experiments}
\subsection{The Learning Process of LtA}
\label{learninglta}
First, we show the learning process of LtA. We consider 3, 5, 8, and 10-objective cases. 

To generate training solution sets $\{S_1,S_2,...,S_L\}$ for each case, we set $L=100$ and $N=100$. That is, we generate 100 solution sets with 100 solutions in each solution set. Solutions in half solution sets (i.e., $\{S_1,S_2,...,S_{50}\}$) are randomly generated on the triangular Pareto front $\sum_{i=1}^m f_i^p=1$, $f_i\geq 0$ for $i=1,...,m$, where $p$ is randomly sampled in $[0.5,2]$ for each solution set. Solutions in the other half solution sets (i.e., $\{S_{51},S_{52},...,S_{100}\}$) are randomly generated on the inverted triangular Pareto front $\sum_{i=1}^m (1-f_i)^p=1$, $0\leq f_i\leq 1$ for $i=1,...,m$, where $p$ is randomly sampled in $[0.5,2]$  for each solution set. In our experiments, $p$ controls the curvature of the Pareto fronts. Thus, 100 solution sets are generated on 100 Pareto fronts with different curvatures (concave, linear, and convex) and shapes (triangular and inverted triangular), which guarantees the training solution sets diverse. 

LtA is applied to the generated 100 solution sets in each case under the following settings. The number of direction vectors is set as 91, 105, 120 and 110 in 3, 5, 8 and 10-objective cases, respectively. The maximum number of iterations of LtA is set as 10,000. In LtA, the reference point for hypervolume calculation is specified as $\mathbf{r} = (1.2, ..., 1.2)$ independent of the number of objectives. We perform 20 independent runs of LtA for each case (i.e., 20 direction vector sets are obtained by LtA for each case).

\begin{figure}[]
\centering                                           
\subfigure[3-objective case]{ 
\includegraphics[scale=0.21]{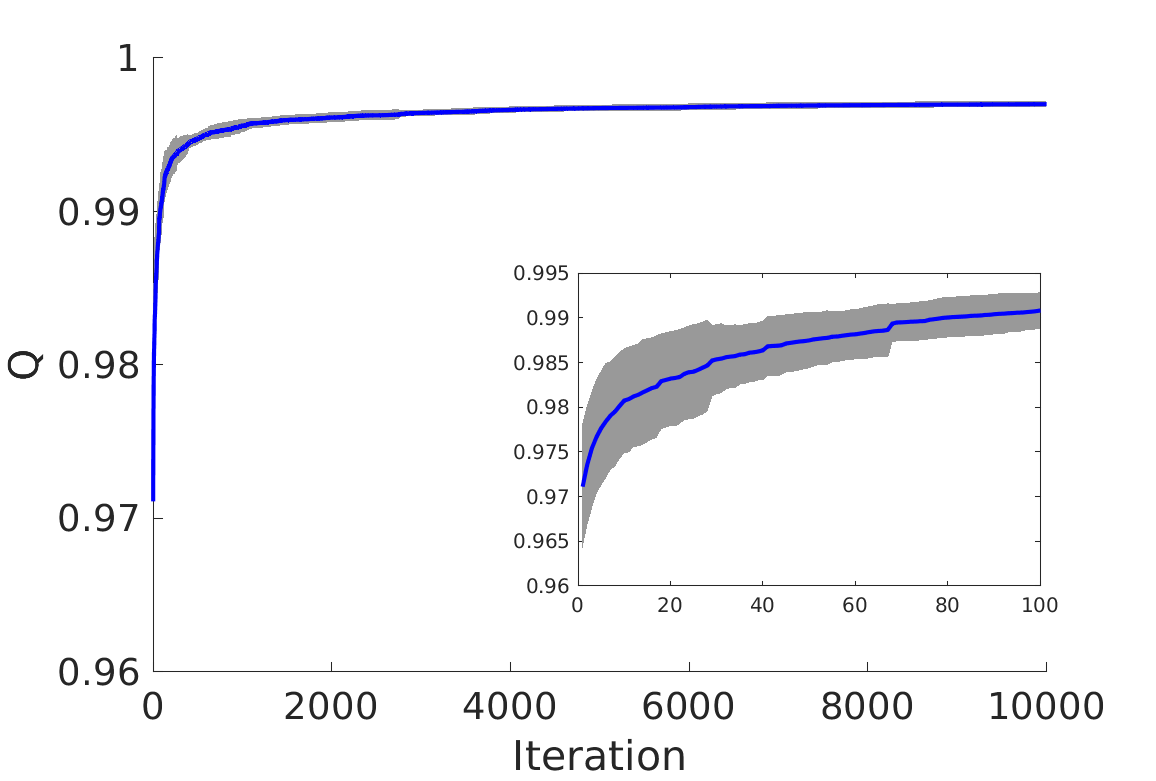}   }            
\subfigure[5-objective case]{ 
\includegraphics[scale=0.21]{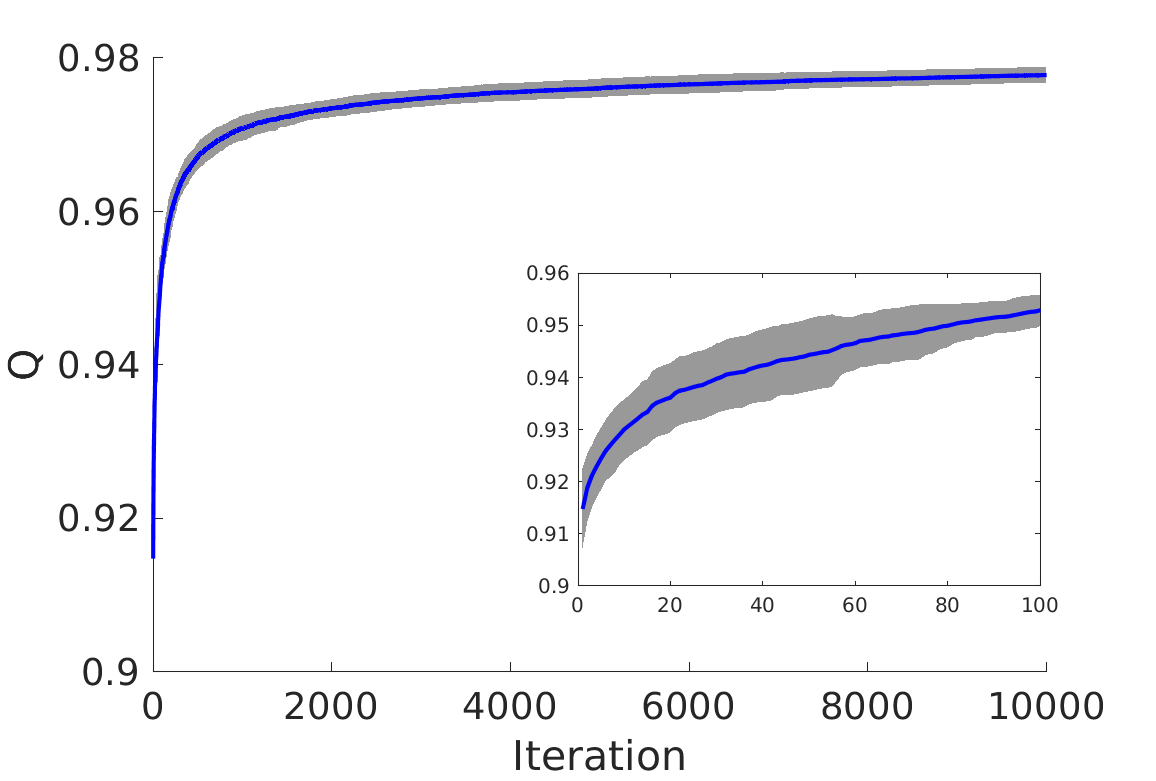} }
\subfigure[8-objective case]{ 
\includegraphics[scale=0.21]{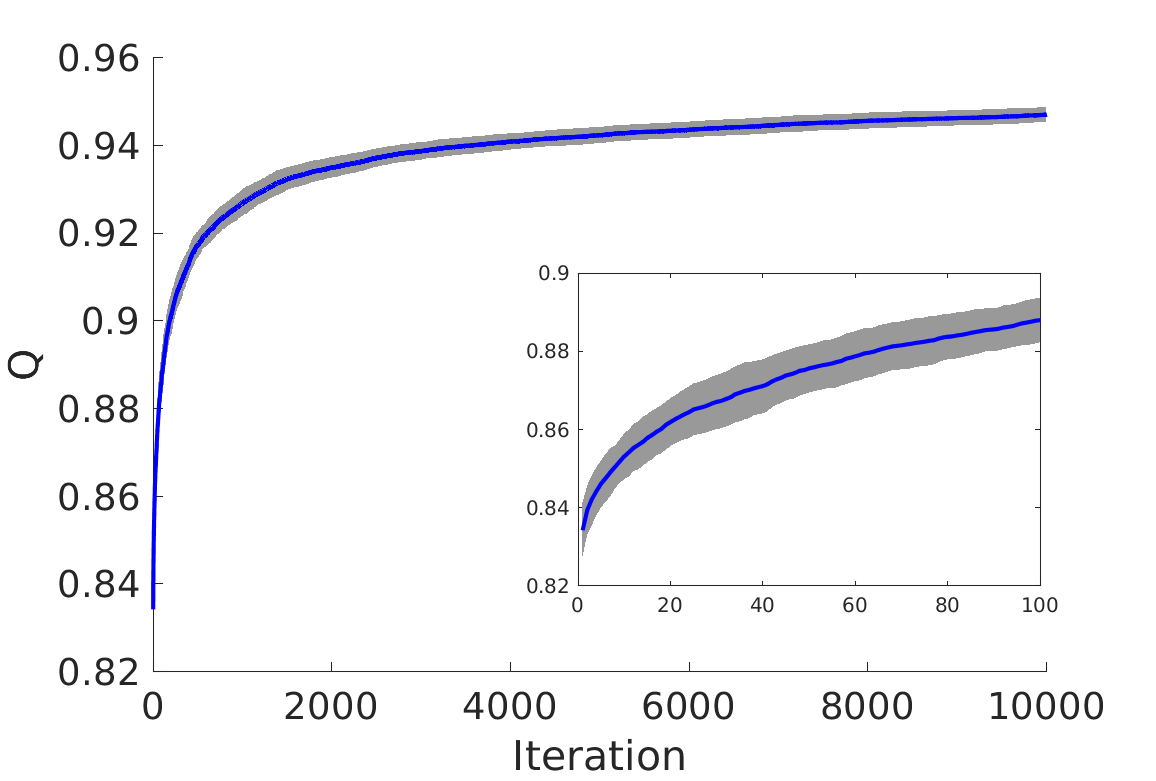} }
\subfigure[10-objective case]{ 
\includegraphics[scale=0.21]{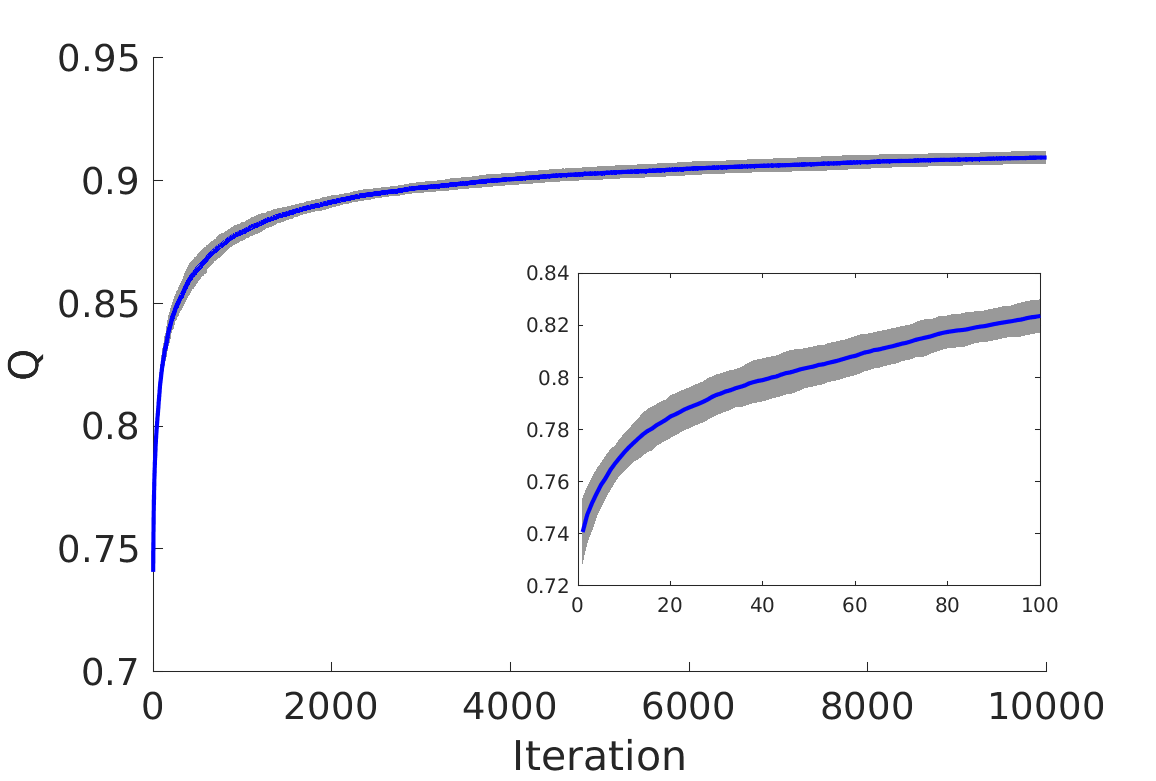} }
\caption{The learning curve of LtA over 20 independent runs in each objective case.} 
\label{learning}                                                        
\end{figure}

Fig. \ref{learning} shows the learning curve of LtA in 3, 5, 8, and 10-objective cases. The blue line is the average $Q$ value over 20 runs, and the grey area denotes the standard deviation. We can see that the $Q$ value is monotonically increasing as the iteration number increases in each case. We can also observe that as the number of objectives increases, the $Q$ value at the end of the learning process decreases. This shows that the learning process becomes more difficult as the number of objectives increases. However, the $Q$ value of the initial direction vector set also decreases with the increase in the number of objectives, and $Q>0.9$ is achieved in all cases. These observations show the success of the learning process of LtA.

\begin{figure}[!htbp]
\centering                                           
\subfigure[3-objective case]{ 
\includegraphics[scale=0.3]{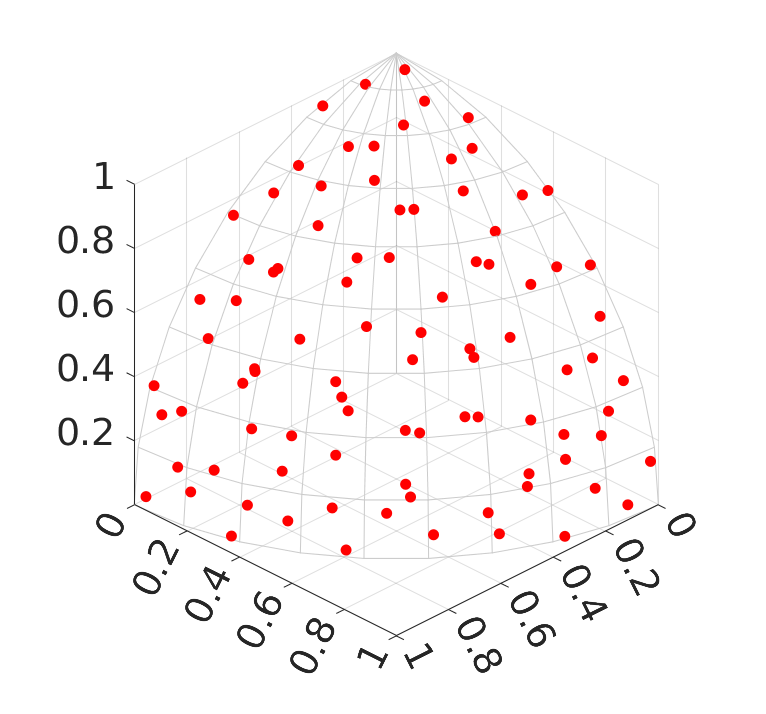}   }            
\subfigure[5-objective case]{ 
\includegraphics[scale=0.21]{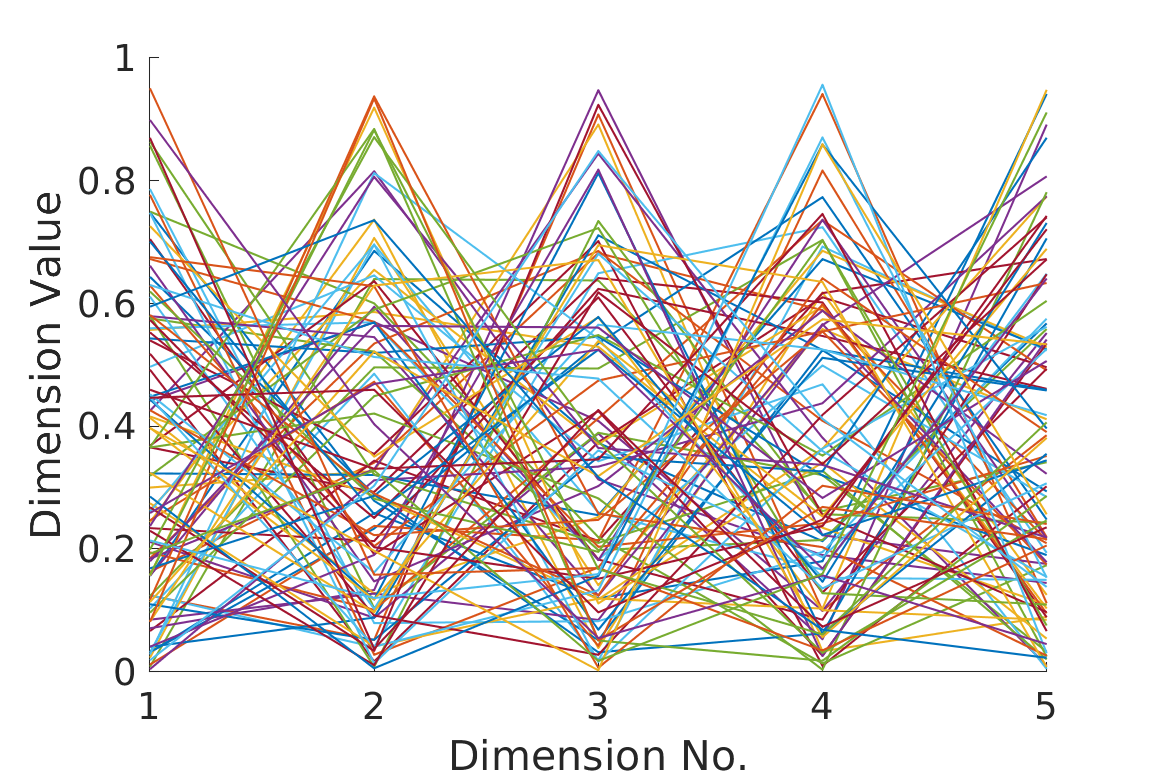} }
\subfigure[8-objective case]{ 
\includegraphics[scale=0.21]{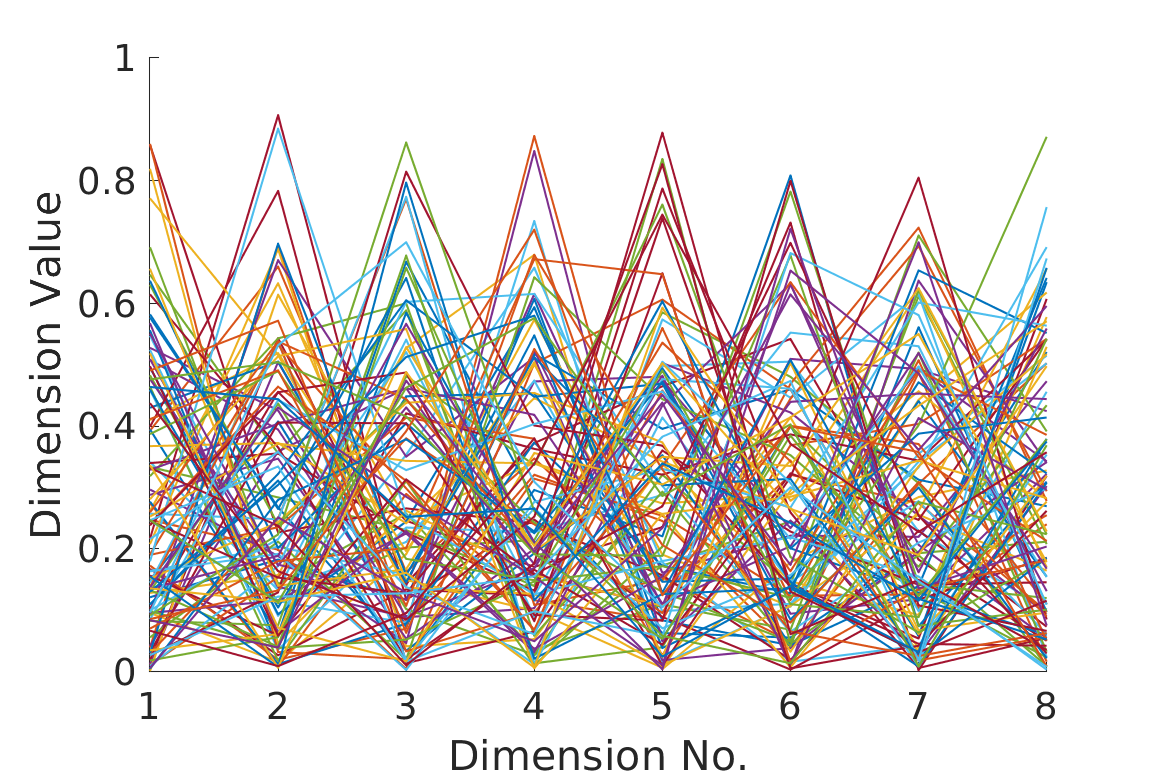} }
\subfigure[10-objective case]{ 
\includegraphics[scale=0.21]{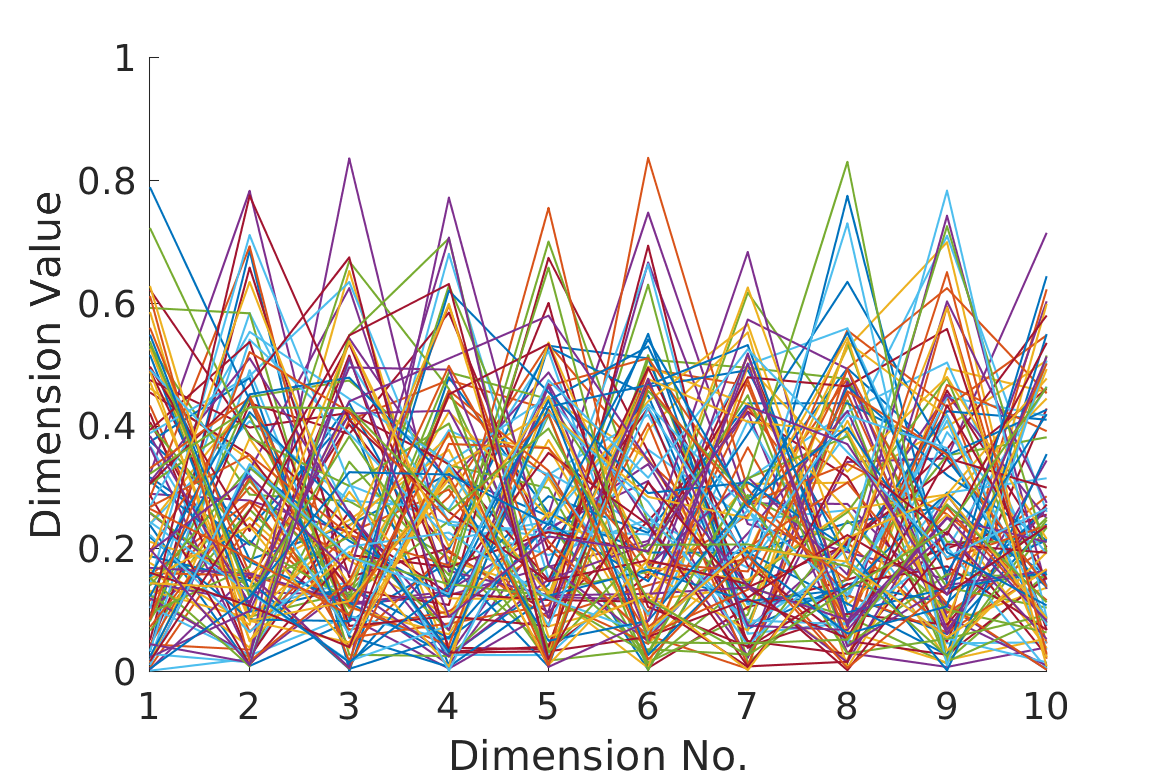} }
\caption{The learned direction vector set in a single run of LtA in each objective case.} 
\label{learnedsets}                                                        
\end{figure}

The learned direction vector set in a single run of LtA for each case is shown in Fig. \ref{learnedsets}. From Fig. \ref{learnedsets} (a) we can visually observe that the learned direction vector set is non-uniform compared with DAS, MSS-D, MSS-U, and Kmeans-U in Fig. \ref{5dvs}.  In the next three subsections, we compare the learned direction vector sets with the direction vector sets generated by the six methods described in Section \ref{generationmethods} through three different experiments.

\begin{table*}[]
 \small \setlength{\tabcolsep}{5pt} 
 \caption{Correct identification rate (CIR) of the $R_2^{\text{HVC}}$ indicator with different direction vector set generation methods on different solution sets.  The number in the parenthesis is the rank of the corresponding method among the seven methods, where a smaller value indicates a better rank.}\centering      \addtolength{\leftskip} {-2cm} \addtolength{\rightskip}{-2cm} 
       \begin{tabular}{|c|c|c|c|c|c|c|c|c|}
 \hline \multicolumn{2}{|c|}{Solution Set}&LtA&DAS&UNV&JAS&MSS-D&MSS-U&Kmeans-U\\ \hline 
\multirow{4}{*}{\makecell{Linear\\Triangular}}&3&\scriptsize{\textbf{81.7\%(1)}}&\scriptsize{62.0\%(6)}&\scriptsize{62.4\%(5)}&\scriptsize{63.7\%(4)}&\scriptsize{67.0\%(3)}&\scriptsize{68.0\%(2)}&\scriptsize{45.0\%(7)}\\\cline{3-9} 
&5&\scriptsize{\textbf{62.9\%(1)}}&\scriptsize{13.0\%(7)}&\scriptsize{49.6\%(2)}&\scriptsize{48.6\%(3)}&\scriptsize{16.0\%(6)}&\scriptsize{18.4\%(5)}&\scriptsize{19.6\%(4)}\\\cline{3-9} 
&8&\scriptsize{\textbf{48.6\%(1)}}&\scriptsize{25.0\%(6)}&\scriptsize{41.0\%(3)}&\scriptsize{35.8\%(4)}&\scriptsize{44.0\%(2)}&\scriptsize{25.8\%(5)}&\scriptsize{9.8\%(7)}\\\cline{3-9} 
&10&\scriptsize{\textbf{49.1\%(1)}}&\scriptsize{13.0\%(7)}&\scriptsize{37.1\%(3)}&\scriptsize{33.0\%(4.5)}&\scriptsize{33.0\%(4.5)}&\scriptsize{45.5\%(2)}&\scriptsize{18.2\%(6)}\\\hline 
\multirow{4}{*}{\makecell{Linear\\Invertedtriangular}}&3&\scriptsize{\textbf{79.2\%(1)}}&\scriptsize{52.0\%(7)}&\scriptsize{63.6\%(3)}&\scriptsize{66.2\%(2)}&\scriptsize{60.0\%(4)}&\scriptsize{59.0\%(5)}&\scriptsize{55.9\%(6)}\\\cline{3-9} 
&5&\scriptsize{\textbf{63.7\%(1)}}&\scriptsize{9.0\%(7)}&\scriptsize{49.5\%(2)}&\scriptsize{46.5\%(3)}&\scriptsize{23.0\%(6)}&\scriptsize{24.1\%(5)}&\scriptsize{25.9\%(4)}\\\cline{3-9} 
&8&\scriptsize{\textbf{50.2\%(1)}}&\scriptsize{1.0\%(6.5)}&\scriptsize{41.9\%(2)}&\scriptsize{24.7\%(4)}&\scriptsize{1.0\%(6.5)}&\scriptsize{6.4\%(5)}&\scriptsize{31.6\%(3)}\\\cline{3-9} 
&10&\scriptsize{28.1\%(3)}&\scriptsize{3.0\%(5)}&\scriptsize{29.3\%(2)}&\scriptsize{5.4\%(4)}&\scriptsize{2.0\%(7)}&\scriptsize{2.5\%(6)}&\scriptsize{\textbf{32.8\%(1)}}\\\hline 
\multirow{4}{*}{\makecell{Concave\\Triangular}}&3&\scriptsize{\textbf{60.0\%(1)}}&\scriptsize{32.0\%(7)}&\scriptsize{51.5\%(3)}&\scriptsize{51.8\%(2)}&\scriptsize{35.0\%(5)}&\scriptsize{34.2\%(6)}&\scriptsize{35.9\%(4)}\\\cline{3-9} 
&5&\scriptsize{\textbf{67.8\%(1)}}&\scriptsize{16.0\%(7)}&\scriptsize{52.2\%(2)}&\scriptsize{49.3\%(3)}&\scriptsize{21.0\%(6)}&\scriptsize{22.7\%(4)}&\scriptsize{22.3\%(5)}\\\cline{3-9} 
&8&\scriptsize{\textbf{69.7\%(1)}}&\scriptsize{54.0\%(6)}&\scriptsize{66.0\%(3)}&\scriptsize{63.5\%(4)}&\scriptsize{55.0\%(5)}&\scriptsize{66.6\%(2)}&\scriptsize{45.4\%(7)}\\\cline{3-9} 
&10&\scriptsize{\textbf{63.5\%(1)}}&\scriptsize{38.0\%(7)}&\scriptsize{57.0\%(4)}&\scriptsize{60.1\%(3)}&\scriptsize{48.0\%(5)}&\scriptsize{62.9\%(2)}&\scriptsize{42.0\%(6)}\\\hline 
\multirow{4}{*}{\makecell{Concave\\Invertedtriangular}}&3&\scriptsize{\textbf{76.0\%(1)}}&\scriptsize{44.0\%(6)}&\scriptsize{58.5\%(3)}&\scriptsize{59.2\%(2)}&\scriptsize{53.0\%(4)}&\scriptsize{51.8\%(5)}&\scriptsize{39.8\%(7)}\\\cline{3-9} 
&5&\scriptsize{\textbf{49.1\%(1)}}&\scriptsize{23.0\%(4)}&\scriptsize{41.3\%(2)}&\scriptsize{39.6\%(3)}&\scriptsize{15.0\%(6)}&\scriptsize{12.0\%(7)}&\scriptsize{16.2\%(5)}\\\cline{3-9} 
&8&\scriptsize{\textbf{58.7\%(1)}}&\scriptsize{0.0\%(7)}&\scriptsize{42.0\%(3)}&\scriptsize{39.8\%(4)}&\scriptsize{12.0\%(6)}&\scriptsize{45.0\%(2)}&\scriptsize{33.9\%(5)}\\\cline{3-9} 
&10&\scriptsize{\textbf{50.9\%(1)}}&\scriptsize{0.0\%(7)}&\scriptsize{36.0\%(4)}&\scriptsize{34.6\%(5)}&\scriptsize{1.0\%(6)}&\scriptsize{38.1\%(3)}&\scriptsize{45.6\%(2)}\\\hline 
\multirow{4}{*}{\makecell{Convex\\Triangular}}&3&\scriptsize{\textbf{71.6\%(1)}}&\scriptsize{47.0\%(6)}&\scriptsize{58.3\%(3)}&\scriptsize{59.5\%(2)}&\scriptsize{49.0\%(5)}&\scriptsize{49.2\%(4)}&\scriptsize{36.2\%(7)}\\\cline{3-9} 
&5&\scriptsize{27.6\%(2)}&\scriptsize{16.0\%(4.5)}&\scriptsize{25.3\%(3)}&\scriptsize{\textbf{29.3\%(1)}}&\scriptsize{16.0\%(4.5)}&\scriptsize{15.9\%(6)}&\scriptsize{3.5\%(7)}\\\cline{3-9} 
&8&\scriptsize{15.6\%(4)}&\scriptsize{35.0\%(2)}&\scriptsize{13.1\%(5)}&\scriptsize{16.9\%(3)}&\scriptsize{\textbf{36.0\%(1)}}&\scriptsize{3.0\%(6)}&\scriptsize{1.1\%(7)}\\\cline{3-9} 
&10&\scriptsize{16.6\%(2)}&\scriptsize{4.0\%(6)}&\scriptsize{11.3\%(4)}&\scriptsize{13.8\%(3)}&\scriptsize{\textbf{43.0\%(1)}}&\scriptsize{4.1\%(5)}&\scriptsize{0.6\%(7)}\\\hline 
\multirow{4}{*}{\makecell{Convex\\Invertedtriangular}}&3&\scriptsize{\textbf{64.8\%(1)}}&\scriptsize{28.0\%(5)}&\scriptsize{48.4\%(3)}&\scriptsize{49.3\%(2)}&\scriptsize{27.0\%(7)}&\scriptsize{27.9\%(6)}&\scriptsize{31.4\%(4)}\\\cline{3-9} 
&5&\scriptsize{89.5\%(2)}&\scriptsize{\textbf{90.0\%(1)}}&\scriptsize{75.0\%(5)}&\scriptsize{77.5\%(4)}&\scriptsize{72.0\%(6)}&\scriptsize{79.9\%(3)}&\scriptsize{37.4\%(7)}\\\cline{3-9} 
&8&\scriptsize{89.2\%(2)}&\scriptsize{\textbf{100.0\%(1)}}&\scriptsize{85.5\%(4)}&\scriptsize{80.5\%(6)}&\scriptsize{73.0\%(7)}&\scriptsize{84.5\%(5)}&\scriptsize{87.3\%(3)}\\\cline{3-9} 
&10&\scriptsize{94.3\%(2)}&\scriptsize{\textbf{100.0\%(1)}}&\scriptsize{88.8\%(4)}&\scriptsize{85.9\%(5)}&\scriptsize{83.0\%(7)}&\scriptsize{84.8\%(6)}&\scriptsize{91.0\%(3)}\\\hline 
\multicolumn{2}{|c|}{Average Rank}&$\mathbf{1.42}$&$5.38$&$3.21$&$3.35$&$5.02$&$4.46$&$5.17$\\ \hline 
\multicolumn{2}{|c|}{Average CIR}&$\mathbf{59.5\%}$&$33.5\%$&$49.4\%$&$47.3\%$&$36.9\%$&$38.8\%$&$33.7\%$\\ \hline 
\end{tabular}
 \label{tab:CIR}\end{table*}
 
\subsection{Experiment 1: Correct Identification Rate}
\label{cirexperiment}
In the first experiment, we use the correct identification rate (CIR) \cite{shang2018r2} to evaluate different direction vector set generation methods. The CIR is a metric which can evaluate the ability of the $R_2^{\text{HVC}}$ indicator to identify the least hypervolume contributor in a solution set. To be more specific, suppose that we have $M$ solution sets $\{T_1,T_2,...,T_M\}$. In each solution set, the true hypervolume contribution of each solution is calculated and the solution with the minimum hypervolume contribution is identified. Using a direction vector set, the approximated hypervolume contribution $R_2^{\text{HVC}}$ of each solution is calculated in each solution set. Then, the solution with the minimum $R_2^{\text{HVC}}$ value is identified. Suppose that the solution with the minimum hypervolume contribution is correctly identified using the approximate calculation for $M'$ out of the $M$ solution sets. Then the CIR value is defined as $M'/M$. A higher value of CIR indicates a better approximation quality of the $R_2^{\text{HVC}}$ indicator.

\subsubsection{Experimental settings}
\label{cirsetting}
We consider six types of Pareto fronts: linear, concave, and convex Pareto fronts with triangular and inverted triangular shapes. For the Pareto fronts with triangular shape, the mathematical expression is $\sum_{i=1}^m f_i^p=1$  and $f_i\geq 0$ for $i=1,...,m$, where $p=1,2,0.5$ correspond to linear, concave, and convex Pareto fronts, respectively. For the Pareto fronts with inverted triangular shape, the mathematical expression is $\sum_{i=1}^m (1-f_i)^p=1$  and $0\leq f_i\leq 1$ for $i=1,...,m$, where $p=1,2,0.5$ correspond to linear, convex, and concave Pareto fronts, respectively. It should be noted that 50 training solution sets were generated from each of triangular and inverted triangular Pareto fronts in the LtA method where the value of $p$ was randomly sampled in $[0.5, 2]$ as described in Section \ref{learninglta}. 

For each type of Pareto front, 100 solutions are randomly sampled on the Pareto front to form a solution set. We generate 100 solution sets for each type of Pareto front. The reference point is set as 1.2 times the nadir point of the Pareto front for hypervolume calculation (i.e., the reference point $\mathbf{r} = (1.2, ..., 1.2)$ is used since the ideal and nadir points are $(0, ..., 0)$ and $(1, ..., 1)$ for all Pareto fronts used in this subsection). 

Each direction vector set generation method is run 20 times on each solution set, and the average CIR is recorded. 

\begin{table*}[]
 \small \setlength{\tabcolsep}{5pt} 
 \caption{Hypervolume performance of the subsets selected by GAHSS with different direction vector set generation methods on different candidate solution sets. The number in the parenthesis is the rank of the corresponding method among the seven methods, where a smaller value indicates a better rank. `$+$', `$-$' and `$\approx$' indicate that the proposed LtA method is `significantly better than', `significantly worse than' and `statistically similar to' the compared method, respectively.
}\centering      \addtolength{\leftskip} {-2cm} \addtolength{\rightskip}{-2cm} 
       \begin{tabular}{|c|c|c|c|c|c|c|c|c|}
 \hline \multicolumn{2}{|c|}{Candidate Solution Set}&LtA&DAS&UNV&JAS&MSS-D&MSS-U&Kmeans-U\\ \hline 
\multirow{4}{*}{\makecell{Linear\\Triangular}}&3&\scriptsize{\textbf{1.5187E+0(1)}}&\scriptsize{1.5162E+0(7,+)}&\scriptsize{1.5182E+0(3,+)}&\scriptsize{1.5179E+0(4,+)}&\scriptsize{1.5170E+0(6,+)}&\scriptsize{1.5175E+0(5,+)}&\scriptsize{1.5184E+0(2,+)}\\\cline{3-9} 
&5&\scriptsize{\textbf{2.4389E+0(1)}}&\scriptsize{2.4305E+0(6,+)}&\scriptsize{2.4354E+0(2,+)}&\scriptsize{2.4333E+0(4,+)}&\scriptsize{2.4302E+0(7,+)}&\scriptsize{2.4321E+0(5,+)}&\scriptsize{2.4345E+0(3,+)}\\\cline{3-9} 
&8&\scriptsize{\textbf{4.2759E+0(1)}}&\scriptsize{4.2709E+0(5,+)}&\scriptsize{4.2684E+0(7,+)}&\scriptsize{4.2719E+0(4,+)}&\scriptsize{4.2757E+0(2,+)}&\scriptsize{4.2729E+0(3,+)}&\scriptsize{4.2695E+0(6,+)}\\\cline{3-9} 
&10&\scriptsize{\textbf{6.1666E+0(1)}}&\scriptsize{6.1598E+0(5,+)}&\scriptsize{6.1549E+0(7,+)}&\scriptsize{6.1610E+0(4,+)}&\scriptsize{6.1571E+0(6,+)}&\scriptsize{6.1648E+0(2,$\approx$)}&\scriptsize{6.1618E+0(3,+)}\\\hline 
\multirow{4}{*}{\makecell{Linear\\Invertedtriangular}}&3&\scriptsize{\textbf{5.3321E-1(1)}}&\scriptsize{5.2964E-1(7,+)}&\scriptsize{5.3193E-1(3,+)}&\scriptsize{5.3084E-1(6,+)}&\scriptsize{5.3189E-1(4,+)}&\scriptsize{5.3160E-1(5,+)}&\scriptsize{5.3314E-1(2,$\approx$)}\\\cline{3-9} 
&5&\scriptsize{7.5481E-2(2)}&\scriptsize{6.8731E-2(6,+)}&\scriptsize{7.5140E-2(3,+)}&\scriptsize{7.3511E-2(4,+)}&\scriptsize{6.7592E-2(7,+)}&\scriptsize{7.1020E-2(5,+)}&\scriptsize{\textbf{7.5701E-2(1,-)}}\\\cline{3-9} 
&8&\scriptsize{1.5130E-3(3)}&\scriptsize{1.1273E-3(7,+)}&\scriptsize{1.5181E-3(2,$\approx$)}&\scriptsize{1.4235E-3(4,+)}&\scriptsize{1.3478E-3(5,+)}&\scriptsize{1.2328E-3(6,+)}&\scriptsize{\textbf{1.5572E-3(1,-)}}\\\cline{3-9} 
&10&\scriptsize{8.6191E-5(2)}&\scriptsize{6.5463E-5(7,+)}&\scriptsize{8.6003E-5(3,$\approx$)}&\scriptsize{7.7188E-5(4,+)}&\scriptsize{6.5496E-5(6,+)}&\scriptsize{7.0402E-5(5,+)}&\scriptsize{\textbf{8.9636E-5(1,-)}}\\\hline 
\multirow{4}{*}{\makecell{Concave\\Triangular}}&3&\scriptsize{\textbf{1.1526E+0(1)}}&\scriptsize{1.1226E+0(7,+)}&\scriptsize{1.1504E+0(3,+)}&\scriptsize{1.1486E+0(4,+)}&\scriptsize{1.1397E+0(6,+)}&\scriptsize{1.1402E+0(5,+)}&\scriptsize{1.1524E+0(2,+)}\\\cline{3-9} 
&5&\scriptsize{\textbf{2.1339E+0(1)}}&\scriptsize{2.0735E+0(7,+)}&\scriptsize{2.1270E+0(3,+)}&\scriptsize{2.1262E+0(4,+)}&\scriptsize{2.0864E+0(6,+)}&\scriptsize{2.1180E+0(5,+)}&\scriptsize{2.1286E+0(2,+)}\\\cline{3-9} 
&8&\scriptsize{\textbf{3.9979E+0(1)}}&\scriptsize{3.9590E+0(6,+)}&\scriptsize{3.9903E+0(5,+)}&\scriptsize{3.9941E+0(2,+)}&\scriptsize{3.9527E+0(7,+)}&\scriptsize{3.9906E+0(4,+)}&\scriptsize{3.9937E+0(3,+)}\\\cline{3-9} 
&10&\scriptsize{5.8599E+0(3)}&\scriptsize{\textbf{5.8678E+0(1,-)}}&\scriptsize{5.8499E+0(6,+)}&\scriptsize{5.8553E+0(5,+)}&\scriptsize{5.8425E+0(7,+)}&\scriptsize{5.8613E+0(2,$\approx$)}&\scriptsize{5.8558E+0(4,+)}\\\hline 
\multirow{4}{*}{\makecell{Concave\\Invertedtriangular}}&3&\scriptsize{\textbf{2.2171E-1(1)}}&\scriptsize{2.0846E-1(5,+)}&\scriptsize{2.2121E-1(3,+)}&\scriptsize{2.2079E-1(4,+)}&\scriptsize{2.0200E-1(6,+)}&\scriptsize{1.9939E-1(7,+)}&\scriptsize{2.2158E-1(2,+)}\\\cline{3-9} 
&5&\scriptsize{\textbf{1.4818E-2(1)}}&\scriptsize{1.1820E-2(7,+)}&\scriptsize{1.4722E-2(3,+)}&\scriptsize{1.4680E-2(4,+)}&\scriptsize{1.2566E-2(6,+)}&\scriptsize{1.3100E-2(5,+)}&\scriptsize{1.4741E-2(2,+)}\\\cline{3-9} 
&8&\scriptsize{\textbf{1.5414E-4(1)}}&\scriptsize{1.2897E-4(7,+)}&\scriptsize{1.5312E-4(2,+)}&\scriptsize{1.5224E-4(4,+)}&\scriptsize{1.3030E-4(6,+)}&\scriptsize{1.4826E-4(5,+)}&\scriptsize{1.5248E-4(3,+)}\\\cline{3-9} 
&10&\scriptsize{\textbf{6.6820E-6(1)}}&\scriptsize{6.5876E-6(4,+)}&\scriptsize{6.5966E-6(3,+)}&\scriptsize{6.5437E-6(5,+)}&\scriptsize{5.9264E-6(7,+)}&\scriptsize{6.4045E-6(6,+)}&\scriptsize{6.6510E-6(2,+)}\\\hline 
\multirow{4}{*}{\makecell{Convex\\Triangular}}&3&\scriptsize{\textbf{1.7104E+0(1)}}&\scriptsize{1.6865E+0(7,+)}&\scriptsize{1.7100E+0(3,+)}&\scriptsize{1.7078E+0(4,+)}&\scriptsize{1.6992E+0(6,+)}&\scriptsize{1.6996E+0(5,+)}&\scriptsize{1.7100E+0(2,+)}\\\cline{3-9} 
&5&\scriptsize{2.4875E+0(3)}&\scriptsize{\textbf{2.4878E+0(1,-)}}&\scriptsize{2.4867E+0(6,+)}&\scriptsize{2.4871E+0(4,+)}&\scriptsize{2.4875E+0(2,$\approx$)}&\scriptsize{2.4870E+0(5,+)}&\scriptsize{2.4863E+0(7,+)}\\\cline{3-9} 
&8&\scriptsize{4.2995E+0(4)}&\scriptsize{4.2901E+0(7,+)}&\scriptsize{4.2995E+0(3,$\approx$)}&\scriptsize{4.2995E+0(2,$\approx$)}&\scriptsize{\textbf{4.2997E+0(1,-)}}&\scriptsize{4.2980E+0(6,+)}&\scriptsize{4.2990E+0(5,+)}\\\cline{3-9} 
&10&\scriptsize{6.1915E+0(2)}&\scriptsize{6.1512E+0(7,+)}&\scriptsize{\textbf{6.1915E+0(1,-)}}&\scriptsize{6.1914E+0(3,$\approx$)}&\scriptsize{6.1877E+0(6,+)}&\scriptsize{6.1897E+0(5,+)}&\scriptsize{6.1911E+0(4,+)}\\\hline 
\multirow{4}{*}{\makecell{Convex\\Invertedtriangular}}&3&\scriptsize{1.0462E+0(2)}&\scriptsize{9.5352E-1(7,+)}&\scriptsize{1.0425E+0(3,+)}&\scriptsize{1.0394E+0(4,+)}&\scriptsize{1.0082E+0(6,+)}&\scriptsize{1.0146E+0(5,+)}&\scriptsize{\textbf{1.0471E+0(1,-)}}\\\cline{3-9} 
&5&\scriptsize{4.4494E-1(2)}&\scriptsize{4.0449E-1(6,+)}&\scriptsize{4.4033E-1(3,+)}&\scriptsize{4.3479E-1(4,+)}&\scriptsize{3.9569E-1(7,+)}&\scriptsize{4.2443E-1(5,+)}&\scriptsize{\textbf{4.4717E-1(1,-)}}\\\cline{3-9} 
&8&\scriptsize{5.3629E-2(2)}&\scriptsize{3.8786E-2(7,+)}&\scriptsize{5.3111E-2(3,+)}&\scriptsize{5.1345E-2(5,+)}&\scriptsize{4.6689E-2(6,+)}&\scriptsize{5.1721E-2(4,+)}&\scriptsize{\textbf{5.4407E-2(1,-)}}\\\cline{3-9} 
&10&\scriptsize{\textbf{9.4647E-3(1)}}&\scriptsize{5.4625E-3(7,+)}&\scriptsize{9.3341E-3(2,+)}&\scriptsize{8.8464E-3(5,+)}&\scriptsize{7.7692E-3(6,+)}&\scriptsize{8.8651E-3(4,+)}&\scriptsize{9.2621E-3(3,+)}\\\hline 
\multicolumn{2}{|c|}{Average Rank}&$\mathbf{1.62}$&$5.96$&$3.42$&$4.04$&$5.58$&$4.75$&$2.62$\\ \hline 
\multicolumn{3}{|c|}{$+$/$-$/$\approx$}&22/2/0&20/1/3&22/0/2&22/1/1&22/0/2&17/6/1\\ \hline 
\end{tabular}
 \label{tab:GAHSS}\end{table*}
 
\subsubsection{Results}
The experimental results are shown in Table \ref{tab:CIR}. We can see that the LtA method achieves the best overall performance. The UNV and JAS methods outperform the DAS, MSS-U, MSS-D, and Kmeans-U methods, which is consistent with the results obtained in \cite{nan2020good}. The counterintuitive observation is that a uniform direction vector set doest not have a good CIR performance. In Table \ref{tab:CIR}, the overall top three methods are LtA, UNV and JAS. As shown in Fig. \ref{5dvs} and Fig. \ref{learnedsets} (a), these three methods generate non-uniform direction vectors (whereas all the other methods generate uniform direction vectors). A possible reason for this counterintuitive observation  is as follows: The hypervolume contribution of a solution has a complex and irregular shape in three and higher dimensional spaces \cite{auger2010theoretically}, which makes the line-based approximation difficult. A uniform direction vector set cannot approximate the hypervolume contribution as expected in such a complex and irregular space. 

\subsection{Experiment 2: GAHSS}
In the second experiment, we use GAHSS\footnote{The source code of GAHSS is available at \url{https: //github.com/HisaoLabSUSTC/GAHSS}.} \cite{shang2021greedy} to compare different direction vector set generation methods. GAHSS is a greedy approximated hypervolume subset selection algorithm for many-objective optimization. GAHSS follows the framework of the standard greedy hypervolume subset selection (GHSS) \cite{bradstreet2007incrementally,guerreiro2016greedy}. In GHSS, a subset is selected from a candidate solution set in a greedy manner. Initially the subset is empty. In each step, the solution in the candidate solution set with the largest hypervolume contribution to the subset is added into the subset. This procedure is repeated until the subset has the pre-specified number of solutions. In GAHSS, the $R_2^{\text{HVC}}$ indicator is used to approximate the hypervolume contribution.

\subsubsection{Experimental settings}
We consider six types of Pareto fronts which are the same as described in Section \ref{cirsetting}. From each of the six Pareto fronts, we randomly sample 10,000 solutions to form a candidate solution set. The number of objectives is set as 3, 5, 8, and 10. As a result, we have 6 (Pareto fronts) $\times$ 4 (specifications of the number of objectives) $=$ 24 candidate solution sets. 

We select 100 solutions from each candidate solution set using GAHSS. In GAHSS, the reference point is set as 1.2 times the nadir point of the candidate solution set. For the hypervolume performance evaluation of the subsets, we use the same reference point specification.

GAHSS equipped with direction vector set generated by each method is performed 20 times on each candidate solution set. Furthermore, the results are analyzed by the Wilcoxon rank sum test with a significance level of 0.05 to determine whether one method shows a statistically significant difference with the other, where `$+$', `$-$' and `$\approx$' indicate that the proposed LtA method is `significantly better than', `significantly worse than' and `statistically similar to' the compared method, respectively.

\subsubsection{Results}
The experimental results are shown in Table \ref{tab:GAHSS}. We can see that the LtA method achieves the best overall performance. The UNV and Kmeans-U methods are worse than the LtA method, but better than the JAS, DAS, MSS-U and MSS-D methods. One observation is that the Kmeans-U method has a poor CIR performance in Table \ref{tab:CIR} but a good hypervolume performance in Table \ref{tab:GAHSS}. This shows that a single experiment cannot fully reflect the performance of a direction vector set generation method. We need to examine the performance of a method from different perspectives. Our proposed LtA method always performs the best in the above two different experiments, which shows its robustness across different tasks.

\subsection{Experiment 3: R2HCA-EMOA}
In the third experiment, we use  R2HCA-EMOA\footnote{The source code of R2HCA-EMOA is available at \url{https://github.com/HisaoLabSUSTC/R2HCA-EMOA}.} \cite{shang2020new} to compare different direction vector set generation methods.  R2HCA-EMOA is a new hypervolume-based EMO algorithm for many-objective optimization. It is similar to SMS-EMOA. The difference is that the  $R_2^{\text{HVC}}$ indicator  is used in R2HCA-EMOA to approximate the hypervolume contribution. In each generation of R2HCA-EMOA, one offspring is generated and added into the population, and one individual with the least $R_2^{\text{HVC}}$ value is removed from the population. In the original R2HCA-EMOA paper, the UNV method is used for direction vector set generation. As reported in \cite{shang2020new}, R2HCA-EMOA achieves better hypervolume performance compared with several state-of-the-art EMO algorithms. In particular, R2HCA-EMOA is better than HypE in terms of both runtime and hypervolume performance.   


\subsubsection{Experimental settings}
We apply R2HCA-EMOA equipped with different direction vector sets to multi/many-objective optimization problems. We choose DTLZ1-4 \cite{deb2005scalable}, WFG1-9 \cite{huband2006review}, and their minus versions (i.e., MinusDTLZ1-4, MinusWFG1-9) \cite{ishibuchi2017performance} as test problems, which are the same as in \cite{shang2020new}. We follow the experimental settings in \cite{shang2020new}, which are listed as follows.
\begin{itemize}
\item The population size is set as 100 for 3, 5, 8, and 10-objective cases.
\item  The maximum number of solution evaluations is set to 100,000 for DTLZ1, DTLZ3, WFG1 and their minus versions, and 30,000 for the other test problems.
\item For the hypervolume calculation in the performance evaluation, we first normalize the obtained solutions by R2HCA-EMOA based on the true Pareto front. Then the reference point is set as $\mathbf{r} = (1.2,...,1.2)$.
\item Each method is run 20 times on each test problem. The results are analyzed by the Wilcoxon rank sum test with a significance level of 0.05 to determine whether one method shows a statistically significant difference with the other, where `$+$', `$-$' and `$\approx$' indicate that the proposed LtA method is `significantly better than', `significantly worse than' and `statistically similar to' the compared method, respectively.
\end{itemize}

\subsubsection{Results}
The experimental results are shown in Tables \ref{tab:DLTZ}-\ref{tab:Minus}. Table \ref{tab:DLTZ} shows the results on DTLZ and WFG test problems, and Table \ref{tab:Minus} shows the results on MinusDTLZ and MinusWFG test problems. From both tables we can see that the LtA method achieves the best overall performance. The UNV method performs the second best. On the contrary, the DAS, MSS-U, and MSS-D are the worst three methods. These three methods also perform badly in the previous two experiments. 

From the above three different experiments, we can see that the LtA method can always achieve the best overall performance, which shows the superiority of the learned direction vector sets over the others for hypervolume contribution approximation.

\begin{table*}[]
 \small \setlength{\tabcolsep}{5pt} 
 \caption{Hypervolume performance of R2HCA-EMOA with different direction vector set generation methods on DTLZ and WFG test problems. The number in the parenthesis is the rank of the corresponding method among the seven methods, where a smaller value indicates a better rank. `$+$', `$-$' and `$\approx$' indicate that the proposed LtA method is `significantly better than', `significantly worse than' and `statistically similar to' the compared method, respectively.
}\centering      \addtolength{\leftskip} {-2cm} \addtolength{\rightskip}{-2cm} 
       \begin{tabular}{|c|c|c|c|c|c|c|c|c|}
 \hline \multicolumn{2}{|c|}{Test Problem}&LtA&DAS&UNV&JAS&MSS-D&MSS-U&Kmeans-U\\ \hline 
\multirow{4}{*}{\makecell{DTLZ1}}&3&\scriptsize{\textbf{1.5202E+0(1)}}&\scriptsize{1.5180E+0(7,+)}&\scriptsize{1.5185E+0(5,+)}&\scriptsize{1.5180E+0(6,+)}&\scriptsize{1.5193E+0(3,+)}&\scriptsize{1.5193E+0(2,+)}&\scriptsize{1.5191E+0(4,+)}\\\cline{3-9} 
&5&\scriptsize{\textbf{2.4448E+0(1)}}&\scriptsize{2.3323E+0(5,+)}&\scriptsize{2.4433E+0(2,+)}&\scriptsize{2.4424E+0(3,+)}&\scriptsize{2.2134E+0(7,+)}&\scriptsize{2.2183E+0(6,+)}&\scriptsize{2.4289E+0(4,+)}\\\cline{3-9} 
&8&\scriptsize{\textbf{4.2921E+0(1)}}&\scriptsize{4.0341E+0(5,+)}&\scriptsize{4.2910E+0(2,+)}&\scriptsize{4.2801E+0(3,+)}&\scriptsize{3.9268E-1(6,+)}&\scriptsize{5.4932E-3(7,+)}&\scriptsize{4.2682E+0(4,+)}\\\cline{3-9} 
&10&\scriptsize{\textbf{6.1881E+0(1)}}&\scriptsize{5.8776E+0(5,+)}&\scriptsize{6.1850E+0(2,+)}&\scriptsize{6.1499E+0(4,+)}&\scriptsize{0(7,+)}&\scriptsize{1.0556E-1(6,+)}&\scriptsize{6.1613E+0(3,+)}\\\hline 
\multirow{4}{*}{\makecell{DTLZ2}}&3&\scriptsize{\textbf{1.1542E+0(1)}}&\scriptsize{1.1470E+0(5,+)}&\scriptsize{1.1524E+0(3,+)}&\scriptsize{1.1516E+0(4,+)}&\scriptsize{8.6359E-1(7,+)}&\scriptsize{9.3806E-1(6,+)}&\scriptsize{1.1535E+0(2,+)}\\\cline{3-9} 
&5&\scriptsize{\textbf{2.1688E+0(1)}}&\scriptsize{1.8968E+0(7,+)}&\scriptsize{2.1639E+0(2,+)}&\scriptsize{2.1631E+0(4,+)}&\scriptsize{1.9663E+0(6,+)}&\scriptsize{1.9898E+0(5,+)}&\scriptsize{2.1635E+0(3,+)}\\\cline{3-9} 
&8&\scriptsize{\textbf{4.1387E+0(1)}}&\scriptsize{7.5664E-1(6,+)}&\scriptsize{4.1332E+0(2,+)}&\scriptsize{4.1320E+0(3,+)}&\scriptsize{7.5602E-1(7,+)}&\scriptsize{1.9956E+0(5,+)}&\scriptsize{4.1314E+0(4,+)}\\\cline{3-9} 
&10&\scriptsize{\textbf{6.0786E+0(1)}}&\scriptsize{1.1461E+0(6,+)}&\scriptsize{6.0237E+0(3,+)}&\scriptsize{5.7487E+0(4,+)}&\scriptsize{1.0970E+0(7,+)}&\scriptsize{1.7787E+0(5,+)}&\scriptsize{6.0723E+0(2,+)}\\\hline 
\multirow{4}{*}{\makecell{DTLZ3}}&3&\scriptsize{\textbf{1.1525E+0(1)}}&\scriptsize{1.1147E+0(5,+)}&\scriptsize{1.1493E+0(3,+)}&\scriptsize{1.1491E+0(4,+)}&\scriptsize{2.9591E-1(6,+)}&\scriptsize{2.8791E-1(7,+)}&\scriptsize{1.1511E+0(2,+)}\\\cline{3-9} 
&5&\scriptsize{\textbf{2.1650E+0(1)}}&\scriptsize{1.8778E+0(5,+)}&\scriptsize{2.1606E+0(3,+)}&\scriptsize{2.1613E+0(2,+)}&\scriptsize{7.6828E-1(6,+)}&\scriptsize{4.4470E-1(7,+)}&\scriptsize{2.1587E+0(4,+)}\\\cline{3-9} 
&8&\scriptsize{\textbf{4.1358E+0(1)}}&\scriptsize{9.0624E-1(5,+)}&\scriptsize{4.1293E+0(2,+)}&\scriptsize{4.1290E+0(3,+)}&\scriptsize{7.7263E-1(6,+)}&\scriptsize{7.5174E-1(7,+)}&\scriptsize{4.1176E+0(4,+)}\\\cline{3-9} 
&10&\scriptsize{\textbf{6.0738E+0(1)}}&\scriptsize{1.1422E+0(5,+)}&\scriptsize{6.0392E+0(3,+)}&\scriptsize{6.0064E+0(4,+)}&\scriptsize{1.1289E+0(6,+)}&\scriptsize{1.0770E+0(7,+)}&\scriptsize{6.0417E+0(2,+)}\\\hline 
\multirow{4}{*}{\makecell{DTLZ4}}&3&\scriptsize{9.1836E-1(6)}&\scriptsize{9.3920E-1(5,-)}&\scriptsize{\textbf{1.0163E+0(1,$\approx$)}}&\scriptsize{9.7867E-1(3,$\approx$)}&\scriptsize{9.7158E-1(4,$\approx$)}&\scriptsize{8.2928E-1(7,+)}&\scriptsize{9.9848E-1(2,$\approx$)}\\\cline{3-9} 
&5&\scriptsize{1.9796E+0(4)}&\scriptsize{1.9563E+0(5,+)}&\scriptsize{2.0058E+0(3,$\approx$)}&\scriptsize{\textbf{2.0835E+0(1,$\approx$)}}&\scriptsize{1.8021E+0(7,+)}&\scriptsize{1.8358E+0(6,$\approx$)}&\scriptsize{2.0565E+0(2,$\approx$)}\\\cline{3-9} 
&8&\scriptsize{4.0440E+0(3)}&\scriptsize{3.5779E+0(6,+)}&\scriptsize{\textbf{4.0611E+0(1,$\approx$)}}&\scriptsize{4.0333E+0(4,$\approx$)}&\scriptsize{3.3983E+0(7,+)}&\scriptsize{3.6873E+0(5,+)}&\scriptsize{4.0525E+0(2,$\approx$)}\\\cline{3-9} 
&10&\scriptsize{5.9428E+0(3)}&\scriptsize{3.5572E+0(7,+)}&\scriptsize{5.9808E+0(2,$\approx$)}&\scriptsize{5.9102E+0(4,$\approx$)}&\scriptsize{4.6288E+0(6,+)}&\scriptsize{5.4198E+0(5,+)}&\scriptsize{\textbf{6.0087E+0(1,-)}}\\\hline 
\multirow{4}{*}{\makecell{WFG1}}&3&\scriptsize{1.6605E+0(5)}&\scriptsize{\textbf{1.6621E+0(1,-)}}&\scriptsize{1.6595E+0(6,+)}&\scriptsize{1.6612E+0(2,$\approx$)}&\scriptsize{1.6606E+0(4,$\approx$)}&\scriptsize{1.6610E+0(3,$\approx$)}&\scriptsize{1.6558E+0(7,+)}\\\cline{3-9} 
&5&\scriptsize{2.4829E+0(2)}&\scriptsize{2.2826E+0(7,+)}&\scriptsize{2.4823E+0(3,$\approx$)}&\scriptsize{\textbf{2.4834E+0(1,$\approx$)}}&\scriptsize{2.3919E+0(6,+)}&\scriptsize{2.4117E+0(5,+)}&\scriptsize{2.4653E+0(4,+)}\\\cline{3-9} 
&8&\scriptsize{\textbf{4.2958E+0(1)}}&\scriptsize{3.7348E+0(6,+)}&\scriptsize{4.2952E+0(3,$\approx$)}&\scriptsize{4.2955E+0(2,$\approx$)}&\scriptsize{3.1224E+0(7,+)}&\scriptsize{4.0284E+0(5,+)}&\scriptsize{4.2686E+0(4,+)}\\\cline{3-9} 
&10&\scriptsize{6.1847E+0(3)}&\scriptsize{5.2248E+0(6,+)}&\scriptsize{\textbf{6.1855E+0(1,$\approx$)}}&\scriptsize{6.1848E+0(2,$\approx$)}&\scriptsize{3.4447E+0(7,+)}&\scriptsize{5.5721E+0(5,+)}&\scriptsize{6.1529E+0(4,+)}\\\hline 
\multirow{4}{*}{\makecell{WFG2}}&3&\scriptsize{\textbf{1.6485E+0(1)}}&\scriptsize{1.6481E+0(2,$\approx$)}&\scriptsize{1.6475E+0(5,$\approx$)}&\scriptsize{1.6465E+0(6,+)}&\scriptsize{1.6481E+0(3,$\approx$)}&\scriptsize{1.6478E+0(4,+)}&\scriptsize{1.6437E+0(7,+)}\\\cline{3-9} 
&5&\scriptsize{\textbf{2.4778E+0(1)}}&\scriptsize{2.0873E+0(7,+)}&\scriptsize{2.4775E+0(2,$\approx$)}&\scriptsize{2.4751E+0(3,$\approx$)}&\scriptsize{2.4073E+0(5,+)}&\scriptsize{2.1659E+0(6,+)}&\scriptsize{2.4488E+0(4,+)}\\\cline{3-9} 
&8&\scriptsize{4.2866E+0(3)}&\scriptsize{9.2025E-1(7,+)}&\scriptsize{4.2901E+0(2,-)}&\scriptsize{\textbf{4.2932E+0(1,-)}}&\scriptsize{2.2111E+0(6,+)}&\scriptsize{3.6559E+0(5,+)}&\scriptsize{4.2568E+0(4,+)}\\\cline{3-9} 
&10&\scriptsize{6.1716E+0(3)}&\scriptsize{1.3759E+0(7,+)}&\scriptsize{6.1784E+0(2,-)}&\scriptsize{\textbf{6.1796E+0(1,-)}}&\scriptsize{2.0010E+0(6,+)}&\scriptsize{5.4928E+0(5,+)}&\scriptsize{6.1368E+0(4,+)}\\\hline 
\multirow{4}{*}{\makecell{WFG3}}&3&\scriptsize{8.3512E-1(2)}&\scriptsize{3.7077E-1(5,+)}&\scriptsize{8.3266E-1(4,$\approx$)}&\scriptsize{8.3329E-1(3,$\approx$)}&\scriptsize{3.3548E-1(6,+)}&\scriptsize{3.2342E-1(7,+)}&\scriptsize{\textbf{8.3639E-1(1,$\approx$)}}\\\cline{3-9} 
&5&\scriptsize{7.3798E-1(3)}&\scriptsize{4.5224E-1(6,+)}&\scriptsize{7.5367E-1(2,$\approx$)}&\scriptsize{\textbf{8.2444E-1(1,-)}}&\scriptsize{4.2108E-1(7,+)}&\scriptsize{4.5855E-1(5,+)}&\scriptsize{5.4354E-1(4,+)}\\\cline{3-9} 
&8&\scriptsize{3.6375E-1(6)}&\scriptsize{7.3466E-1(2,-)}&\scriptsize{5.3724E-1(5,$\approx$)}&\scriptsize{\textbf{9.9649E-1(1,-)}}&\scriptsize{6.4246E-1(4,-)}&\scriptsize{6.8834E-1(3,-)}&\scriptsize{3.5942E-3(7,+)}\\\cline{3-9} 
&10&\scriptsize{3.1013E-1(6)}&\scriptsize{6.0244E-1(4,-)}&\scriptsize{5.3879E-1(5,$\approx$)}&\scriptsize{\textbf{1.0251E+0(1,-)}}&\scriptsize{6.0961E-1(3,-)}&\scriptsize{7.0591E-1(2,-)}&\scriptsize{0(7,+)}\\\hline 
\multirow{4}{*}{\makecell{WFG4}}&3&\scriptsize{\textbf{1.1515E+0(1)}}&\scriptsize{1.1462E+0(7,+)}&\scriptsize{1.1494E+0(3,+)}&\scriptsize{1.1485E+0(5,+)}&\scriptsize{1.1484E+0(6,+)}&\scriptsize{1.1491E+0(4,+)}&\scriptsize{1.1510E+0(2,$\approx$)}\\\cline{3-9} 
&5&\scriptsize{\textbf{2.1558E+0(1)}}&\scriptsize{7.0988E-1(7,+)}&\scriptsize{2.1507E+0(2,+)}&\scriptsize{2.1502E+0(3,+)}&\scriptsize{1.5675E+0(5,+)}&\scriptsize{1.4664E+0(6,+)}&\scriptsize{2.1430E+0(4,+)}\\\cline{3-9} 
&8&\scriptsize{\textbf{4.0940E+0(1)}}&\scriptsize{9.4224E-1(7,+)}&\scriptsize{3.9727E+0(3,+)}&\scriptsize{3.8740E+0(4,$\approx$)}&\scriptsize{1.0861E+0(6,+)}&\scriptsize{2.0752E+0(5,+)}&\scriptsize{4.0642E+0(2,+)}\\\cline{3-9} 
&10&\scriptsize{5.6362E+0(3)}&\scriptsize{1.4523E+0(6,+)}&\scriptsize{5.6606E+0(2,$\approx$)}&\scriptsize{5.1627E+0(4,+)}&\scriptsize{1.3888E+0(7,+)}&\scriptsize{3.1301E+0(5,+)}&\scriptsize{\textbf{5.9533E+0(1,-)}}\\\hline 
\multirow{4}{*}{\makecell{WFG5}}&3&\scriptsize{\textbf{1.0886E+0(1)}}&\scriptsize{1.0711E+0(5,+)}&\scriptsize{1.0867E+0(3,+)}&\scriptsize{1.0859E+0(4,+)}&\scriptsize{1.0291E+0(7,+)}&\scriptsize{1.0479E+0(6,+)}&\scriptsize{1.0878E+0(2,+)}\\\cline{3-9} 
&5&\scriptsize{\textbf{2.0497E+0(1)}}&\scriptsize{4.1498E-1(7,+)}&\scriptsize{2.0440E+0(2,+)}&\scriptsize{2.0438E+0(3,+)}&\scriptsize{1.8743E+0(6,+)}&\scriptsize{1.8898E+0(5,+)}&\scriptsize{2.0413E+0(4,+)}\\\cline{3-9} 
&8&\scriptsize{\textbf{3.8894E+0(1)}}&\scriptsize{6.4009E-1(6,+)}&\scriptsize{3.8801E+0(2,+)}&\scriptsize{3.8614E+0(4,+)}&\scriptsize{6.3568E-1(7,+)}&\scriptsize{3.1893E+0(5,+)}&\scriptsize{3.8656E+0(3,+)}\\\cline{3-9} 
&10&\scriptsize{5.6157E+0(2)}&\scriptsize{9.6833E-1(6,+)}&\scriptsize{5.5340E+0(3,+)}&\scriptsize{5.0287E+0(4,+)}&\scriptsize{9.0662E-1(7,+)}&\scriptsize{2.1315E+0(5,+)}&\scriptsize{\textbf{5.6547E+0(1,-)}}\\\hline 
\multirow{4}{*}{\makecell{WFG6}}&3&\scriptsize{1.0695E+0(2)}&\scriptsize{1.0563E+0(7,+)}&\scriptsize{\textbf{1.0709E+0(1,$\approx$)}}&\scriptsize{1.0649E+0(4,$\approx$)}&\scriptsize{1.0592E+0(6,$\approx$)}&\scriptsize{1.0606E+0(5,$\approx$)}&\scriptsize{1.0660E+0(3,$\approx$)}\\\cline{3-9} 
&5&\scriptsize{2.0037E+0(3)}&\scriptsize{1.2638E+0(7,+)}&\scriptsize{\textbf{2.0242E+0(1,$\approx$)}}&\scriptsize{2.0117E+0(2,$\approx$)}&\scriptsize{1.9050E+0(5,+)}&\scriptsize{1.8943E+0(6,+)}&\scriptsize{1.9937E+0(4,$\approx$)}\\\cline{3-9} 
&8&\scriptsize{\textbf{3.8387E+0(1)}}&\scriptsize{1.0264E+0(7,+)}&\scriptsize{3.7933E+0(2,+)}&\scriptsize{3.7647E+0(3,$\approx$)}&\scriptsize{1.4593E+0(6,+)}&\scriptsize{3.5258E+0(5,+)}&\scriptsize{3.7639E+0(4,+)}\\\cline{3-9} 
&10&\scriptsize{\textbf{5.5876E+0(1)}}&\scriptsize{1.4629E+0(7,+)}&\scriptsize{5.5282E+0(2,$\approx$)}&\scriptsize{5.1685E+0(4,+)}&\scriptsize{1.4972E+0(6,+)}&\scriptsize{3.9992E+0(5,+)}&\scriptsize{5.5104E+0(3,+)}\\\hline 
\multirow{4}{*}{\makecell{WFG7}}&3&\scriptsize{\textbf{1.1538E+0(1)}}&\scriptsize{1.1451E+0(7,+)}&\scriptsize{1.1512E+0(3,+)}&\scriptsize{1.1502E+0(4,+)}&\scriptsize{1.1467E+0(6,+)}&\scriptsize{1.1469E+0(5,+)}&\scriptsize{1.1529E+0(2,+)}\\\cline{3-9} 
&5&\scriptsize{\textbf{2.1652E+0(1)}}&\scriptsize{9.9300E-1(7,+)}&\scriptsize{2.1587E+0(2,+)}&\scriptsize{2.1583E+0(3,+)}&\scriptsize{1.9551E+0(6,+)}&\scriptsize{2.0436E+0(5,+)}&\scriptsize{2.1573E+0(4,+)}\\\cline{3-9} 
&8&\scriptsize{\textbf{4.1252E+0(1)}}&\scriptsize{1.0041E+0(7,+)}&\scriptsize{4.1196E+0(2,+)}&\scriptsize{4.0383E+0(4,+)}&\scriptsize{1.2485E+0(6,+)}&\scriptsize{3.3479E+0(5,+)}&\scriptsize{4.1037E+0(3,+)}\\\cline{3-9} 
&10&\scriptsize{5.9645E+0(2)}&\scriptsize{1.5584E+0(7,+)}&\scriptsize{5.9248E+0(3,$\approx$)}&\scriptsize{5.5189E+0(4,+)}&\scriptsize{1.6144E+0(6,+)}&\scriptsize{3.6273E+0(5,+)}&\scriptsize{\textbf{6.0331E+0(1,$\approx$)}}\\\hline 
\multirow{4}{*}{\makecell{WFG8}}&3&\scriptsize{\textbf{1.0371E+0(1)}}&\scriptsize{1.0177E+0(7,+)}&\scriptsize{1.0343E+0(3,+)}&\scriptsize{1.0342E+0(4,+)}&\scriptsize{1.0201E+0(5,+)}&\scriptsize{1.0182E+0(6,+)}&\scriptsize{1.0352E+0(2,+)}\\\cline{3-9} 
&5&\scriptsize{\textbf{1.9549E+0(1)}}&\scriptsize{4.1972E-1(7,+)}&\scriptsize{1.9388E+0(3,+)}&\scriptsize{1.9398E+0(2,+)}&\scriptsize{7.4087E-1(6,+)}&\scriptsize{1.6766E+0(5,+)}&\scriptsize{1.9293E+0(4,+)}\\\cline{3-9} 
&8&\scriptsize{\textbf{3.7663E+0(1)}}&\scriptsize{6.7556E-1(7,+)}&\scriptsize{3.7333E+0(2,$\approx$)}&\scriptsize{3.6088E+0(4,+)}&\scriptsize{6.7785E-1(6,+)}&\scriptsize{2.2314E+0(5,+)}&\scriptsize{3.7049E+0(3,+)}\\\cline{3-9} 
&10&\scriptsize{5.4588E+0(2)}&\scriptsize{1.0395E+0(6,+)}&\scriptsize{5.3833E+0(3,$\approx$)}&\scriptsize{5.0956E+0(4,+)}&\scriptsize{9.9181E-1(7,+)}&\scriptsize{3.9871E+0(5,+)}&\scriptsize{\textbf{5.5474E+0(1,-)}}\\\hline 
\multirow{4}{*}{\makecell{WFG9}}&3&\scriptsize{\textbf{1.1298E+0(1)}}&\scriptsize{1.1259E+0(5,+)}&\scriptsize{1.1275E+0(4,$\approx$)}&\scriptsize{1.1275E+0(3,$\approx$)}&\scriptsize{1.0949E+0(7,+)}&\scriptsize{1.1015E+0(6,+)}&\scriptsize{1.1279E+0(2,$\approx$)}\\\cline{3-9} 
&5&\scriptsize{\textbf{2.0954E+0(1)}}&\scriptsize{3.8072E-1(7,+)}&\scriptsize{2.0833E+0(3,+)}&\scriptsize{2.0879E+0(2,+)}&\scriptsize{1.8499E+0(6,+)}&\scriptsize{1.9475E+0(5,+)}&\scriptsize{2.0624E+0(4,+)}\\\cline{3-9} 
&8&\scriptsize{\textbf{3.9524E+0(1)}}&\scriptsize{7.0503E-1(6,+)}&\scriptsize{3.9341E+0(2,$\approx$)}&\scriptsize{3.8575E+0(3,+)}&\scriptsize{6.5629E-1(7,+)}&\scriptsize{2.6976E+0(5,+)}&\scriptsize{3.7956E+0(4,+)}\\\cline{3-9} 
&10&\scriptsize{\textbf{5.5980E+0(1)}}&\scriptsize{9.2488E-1(6,+)}&\scriptsize{5.5309E+0(2,$\approx$)}&\scriptsize{4.7906E+0(4,+)}&\scriptsize{9.0293E-1(7,+)}&\scriptsize{1.8048E+0(5,+)}&\scriptsize{5.4972E+0(3,+)}\\\hline 
\multicolumn{2}{|c|}{Average Rank}&$\mathbf{1.85}$&$5.90$&$2.63$&$3.17$&$5.98$&$5.23$&$3.23$\\ \hline 
\multicolumn{3}{|c|}{$+$/$-$/$\approx$}&47/4/1&27/2/23&32/5/15&46/2/4&47/2/3&39/4/9\\ \hline 
\end{tabular}
 \label{tab:DLTZ}\end{table*}

\begin{table*}[]
 \small \setlength{\tabcolsep}{5pt} 
 \caption{Hypervolume performance of R2HCA-EMOA with different direction vector set generation methods on MinusDTLZ and MinusWFG test problems. The number in the parenthesis is the rank of the corresponding method among the seven methods, where a smaller value indicates a better rank. `$+$', `$-$' and `$\approx$' indicate that the proposed LtA method is `significantly better than', `significantly worse than' and `statistically similar to' the compared method, respectively.
}\centering      \addtolength{\leftskip} {-2cm} \addtolength{\rightskip}{-2cm} 
       \begin{tabular}{|c|c|c|c|c|c|c|c|c|}
 \hline \multicolumn{2}{|c|}{Test Problem}&LtA&DAS&UNV&JAS&MSS-D&MSS-U&Kmeans-U\\ \hline 
\multirow{4}{*}{\makecell{Minus\\DTLZ1}}&3&\scriptsize{\textbf{5.3281E-1(1)}}&\scriptsize{5.2892E-1(7,+)}&\scriptsize{5.3018E-1(5,+)}&\scriptsize{5.2957E-1(6,+)}&\scriptsize{5.3115E-1(3,+)}&\scriptsize{5.3110E-1(4,+)}&\scriptsize{5.3186E-1(2,+)}\\\cline{3-9} 
&5&\scriptsize{\textbf{7.6669E-2(1)}}&\scriptsize{6.8284E-2(7,+)}&\scriptsize{7.5840E-2(2,+)}&\scriptsize{7.5412E-2(4,+)}&\scriptsize{7.2006E-2(6,+)}&\scriptsize{7.3846E-2(5,+)}&\scriptsize{7.5573E-2(3,+)}\\\cline{3-9} 
&8&\scriptsize{1.3309E-3(4)}&\scriptsize{7.3745E-4(7,+)}&\scriptsize{1.3391E-3(3,$\approx$)}&\scriptsize{1.4275E-3(2,-)}&\scriptsize{1.0659E-3(6,+)}&\scriptsize{\textbf{1.4359E-3(1,-)}}&\scriptsize{1.2697E-3(5,+)}\\\cline{3-9} 
&10&\scriptsize{7.2624E-5(4)}&\scriptsize{3.6615E-5(7,+)}&\scriptsize{7.5218E-5(3,-)}&\scriptsize{\textbf{8.3475E-5(1,-)}}&\scriptsize{3.7771E-5(6,+)}&\scriptsize{7.7449E-5(2,-)}&\scriptsize{6.7243E-5(5,+)}\\\hline 
\multirow{4}{*}{\makecell{Minus\\DTLZ2}}&3&\scriptsize{1.0223E+0(3)}&\scriptsize{1.0153E+0(7,+)}&\scriptsize{\textbf{1.0241E+0(1,$\approx$)}}&\scriptsize{1.0197E+0(5,$\approx$)}&\scriptsize{1.0232E+0(2,$\approx$)}&\scriptsize{1.0200E+0(4,$\approx$)}&\scriptsize{1.0194E+0(6,$\approx$)}\\\cline{3-9} 
&5&\scriptsize{\textbf{4.3677E-1(1)}}&\scriptsize{3.8126E-1(7,+)}&\scriptsize{4.3234E-1(2,+)}&\scriptsize{4.2485E-1(4,+)}&\scriptsize{3.9953E-1(6,+)}&\scriptsize{4.1714E-1(5,+)}&\scriptsize{4.2791E-1(3,+)}\\\cline{3-9} 
&8&\scriptsize{\textbf{5.3951E-2(1)}}&\scriptsize{2.6631E-2(7,+)}&\scriptsize{5.2565E-2(3,+)}&\scriptsize{5.2781E-2(2,+)}&\scriptsize{3.1766E-2(6,+)}&\scriptsize{5.0057E-2(5,+)}&\scriptsize{5.0279E-2(4,+)}\\\cline{3-9} 
&10&\scriptsize{\textbf{9.8444E-3(1)}}&\scriptsize{4.5284E-3(6,+)}&\scriptsize{9.5344E-3(2,+)}&\scriptsize{9.1182E-3(3,+)}&\scriptsize{3.8985E-3(7,+)}&\scriptsize{7.7136E-3(5,+)}&\scriptsize{9.0846E-3(4,+)}\\\hline 
\multirow{4}{*}{\makecell{Minus\\DTLZ3}}&3&\scriptsize{1.0226E+0(2)}&\scriptsize{1.0164E+0(7,+)}&\scriptsize{1.0213E+0(3,$\approx$)}&\scriptsize{1.0197E+0(6,$\approx$)}&\scriptsize{\textbf{1.0228E+0(1,$\approx$)}}&\scriptsize{1.0203E+0(5,$\approx$)}&\scriptsize{1.0211E+0(4,$\approx$)}\\\cline{3-9} 
&5&\scriptsize{\textbf{4.3815E-1(1)}}&\scriptsize{3.7919E-1(7,+)}&\scriptsize{4.3365E-1(2,+)}&\scriptsize{4.2519E-1(4,+)}&\scriptsize{4.0196E-1(6,+)}&\scriptsize{4.1809E-1(5,+)}&\scriptsize{4.2646E-1(3,+)}\\\cline{3-9} 
&8&\scriptsize{\textbf{5.4059E-2(1)}}&\scriptsize{2.7226E-2(7,+)}&\scriptsize{5.2562E-2(3,+)}&\scriptsize{5.2731E-2(2,+)}&\scriptsize{3.1793E-2(6,+)}&\scriptsize{4.9848E-2(5,+)}&\scriptsize{5.0156E-2(4,+)}\\\cline{3-9} 
&10&\scriptsize{\textbf{9.8291E-3(1)}}&\scriptsize{4.3990E-3(6,+)}&\scriptsize{9.5351E-3(2,+)}&\scriptsize{9.1063E-3(3,+)}&\scriptsize{3.8725E-3(7,+)}&\scriptsize{7.6838E-3(5,+)}&\scriptsize{8.9476E-3(4,+)}\\\hline 
\multirow{4}{*}{\makecell{Minus\\DTLZ4}}&3&\scriptsize{1.0367E+0(2)}&\scriptsize{1.0338E+0(7,+)}&\scriptsize{1.0357E+0(3,$\approx$)}&\scriptsize{1.0342E+0(5,$\approx$)}&\scriptsize{1.0340E+0(6,+)}&\scriptsize{1.0342E+0(4,+)}&\scriptsize{\textbf{1.0370E+0(1,$\approx$)}}\\\cline{3-9} 
&5&\scriptsize{\textbf{4.4206E-1(1)}}&\scriptsize{3.9366E-1(7,+)}&\scriptsize{4.3956E-1(2,+)}&\scriptsize{4.3541E-1(4,+)}&\scriptsize{4.1554E-1(6,+)}&\scriptsize{4.2873E-1(5,+)}&\scriptsize{4.3637E-1(3,+)}\\\cline{3-9} 
&8&\scriptsize{\textbf{5.1881E-2(1)}}&\scriptsize{3.0133E-2(7,+)}&\scriptsize{4.9873E-2(4,+)}&\scriptsize{5.1746E-2(3,$\approx$)}&\scriptsize{3.8244E-2(6,+)}&\scriptsize{5.1821E-2(2,$\approx$)}&\scriptsize{4.6407E-2(5,+)}\\\cline{3-9} 
&10&\scriptsize{\textbf{9.4404E-3(1)}}&\scriptsize{5.1662E-3(6,+)}&\scriptsize{9.0095E-3(3,+)}&\scriptsize{9.1009E-3(2,+)}&\scriptsize{4.7429E-3(7,+)}&\scriptsize{8.9090E-3(4,+)}&\scriptsize{7.9750E-3(5,+)}\\\hline 
\multirow{4}{*}{\makecell{Minus\\WFG1}}&3&\scriptsize{3.3423E-1(5)}&\scriptsize{3.3429E-1(4,$\approx$)}&\scriptsize{3.3448E-1(3,$\approx$)}&\scriptsize{3.3357E-1(6,+)}&\scriptsize{\textbf{3.3552E-1(1,-)}}&\scriptsize{3.3540E-1(2,$\approx$)}&\scriptsize{3.3339E-1(7,+)}\\\cline{3-9} 
&5&\scriptsize{\textbf{3.1940E-2(1)}}&\scriptsize{1.7595E-2(7,+)}&\scriptsize{3.1656E-2(2,+)}&\scriptsize{3.1462E-2(3,+)}&\scriptsize{1.7725E-2(6,+)}&\scriptsize{2.0174E-2(5,+)}&\scriptsize{3.0709E-2(4,+)}\\\cline{3-9} 
&8&\scriptsize{\textbf{4.4831E-4(1)}}&\scriptsize{1.0366E-4(7,+)}&\scriptsize{4.4422E-4(2,$\approx$)}&\scriptsize{4.4064E-4(3,+)}&\scriptsize{1.3768E-4(5,+)}&\scriptsize{1.1516E-4(6,+)}&\scriptsize{4.3740E-4(4,+)}\\\cline{3-9} 
&10&\scriptsize{\textbf{2.2073E-5(1)}}&\scriptsize{3.3994E-6(7,+)}&\scriptsize{2.1676E-5(2,+)}&\scriptsize{2.1673E-5(3,+)}&\scriptsize{5.3028E-6(5,+)}&\scriptsize{4.5871E-6(6,+)}&\scriptsize{2.1403E-5(4,+)}\\\hline 
\multirow{4}{*}{\makecell{Minus\\WFG2}}&3&\scriptsize{\textbf{6.9566E-1(1)}}&\scriptsize{6.0212E-1(7,+)}&\scriptsize{6.9326E-1(3,+)}&\scriptsize{6.9046E-1(4,+)}&\scriptsize{6.0348E-1(6,+)}&\scriptsize{6.0405E-1(5,+)}&\scriptsize{6.9450E-1(2,+)}\\\cline{3-9} 
&5&\scriptsize{\textbf{7.8748E-2(1)}}&\scriptsize{4.5479E-2(7,+)}&\scriptsize{7.7269E-2(3,+)}&\scriptsize{7.5484E-2(4,+)}&\scriptsize{5.3256E-2(5,+)}&\scriptsize{5.2642E-2(6,+)}&\scriptsize{7.7590E-2(2,+)}\\\cline{3-9} 
&8&\scriptsize{1.2964E-3(2)}&\scriptsize{2.3660E-4(7,+)}&\scriptsize{1.2744E-3(3,+)}&\scriptsize{1.2126E-3(4,+)}&\scriptsize{4.6636E-4(6,+)}&\scriptsize{4.7845E-4(5,+)}&\scriptsize{\textbf{1.3301E-3(1,-)}}\\\cline{3-9} 
&10&\scriptsize{7.0343E-5(2)}&\scriptsize{1.7874E-5(6,+)}&\scriptsize{6.7811E-5(3,+)}&\scriptsize{6.1146E-5(4,+)}&\scriptsize{7.5689E-6(7,+)}&\scriptsize{2.3313E-5(5,+)}&\scriptsize{\textbf{7.2826E-5(1,-)}}\\\hline 
\multirow{4}{*}{\makecell{Minus\\WFG3}}&3&\scriptsize{\textbf{5.3268E-1(1)}}&\scriptsize{5.2817E-1(7,+)}&\scriptsize{5.2993E-1(5,+)}&\scriptsize{5.2948E-1(6,+)}&\scriptsize{5.3021E-1(4,+)}&\scriptsize{5.3072E-1(3,+)}&\scriptsize{5.3196E-1(2,+)}\\\cline{3-9} 
&5&\scriptsize{\textbf{7.6306E-2(1)}}&\scriptsize{6.7232E-2(7,+)}&\scriptsize{7.5347E-2(2,+)}&\scriptsize{7.4894E-2(4,+)}&\scriptsize{7.0653E-2(6,+)}&\scriptsize{7.3274E-2(5,+)}&\scriptsize{7.5208E-2(3,+)}\\\cline{3-9} 
&8&\scriptsize{1.3085E-3(4)}&\scriptsize{6.3092E-4(7,+)}&\scriptsize{1.3092E-3(3,$\approx$)}&\scriptsize{1.3915E-3(2,-)}&\scriptsize{9.7067E-4(6,+)}&\scriptsize{\textbf{1.4004E-3(1,-)}}&\scriptsize{1.2499E-3(5,+)}\\\cline{3-9} 
&10&\scriptsize{7.0906E-5(4)}&\scriptsize{3.0285E-5(7,+)}&\scriptsize{7.3364E-5(3,-)}&\scriptsize{\textbf{8.0643E-5(1,-)}}&\scriptsize{3.1368E-5(6,+)}&\scriptsize{7.5017E-5(2,-)}&\scriptsize{6.6267E-5(5,+)}\\\hline 
\multirow{4}{*}{\makecell{Minus\\WFG4}}&3&\scriptsize{\textbf{1.0321E+0(1)}}&\scriptsize{1.0259E+0(6,+)}&\scriptsize{1.0320E+0(2,$\approx$)}&\scriptsize{1.0282E+0(5,+)}&\scriptsize{1.0291E+0(4,+)}&\scriptsize{1.0255E+0(7,+)}&\scriptsize{1.0318E+0(3,$\approx$)}\\\cline{3-9} 
&5&\scriptsize{\textbf{4.4116E-1(1)}}&\scriptsize{3.9211E-1(7,+)}&\scriptsize{4.3892E-1(2,+)}&\scriptsize{4.3218E-1(4,+)}&\scriptsize{4.0946E-1(6,+)}&\scriptsize{4.2382E-1(5,+)}&\scriptsize{4.3558E-1(3,+)}\\\cline{3-9} 
&8&\scriptsize{\textbf{5.2711E-2(1)}}&\scriptsize{2.8929E-2(7,+)}&\scriptsize{5.1156E-2(4,+)}&\scriptsize{5.2474E-2(2,$\approx$)}&\scriptsize{3.5818E-2(6,+)}&\scriptsize{5.1316E-2(3,+)}&\scriptsize{4.8278E-2(5,+)}\\\cline{3-9} 
&10&\scriptsize{\textbf{9.7403E-3(1)}}&\scriptsize{5.3507E-3(6,+)}&\scriptsize{9.3870E-3(2,+)}&\scriptsize{9.2609E-3(3,+)}&\scriptsize{4.5647E-3(7,+)}&\scriptsize{8.4873E-3(4,+)}&\scriptsize{8.4114E-3(5,+)}\\\hline 
\multirow{4}{*}{\makecell{Minus\\WFG5}}&3&\scriptsize{\textbf{1.0274E+0(1)}}&\scriptsize{1.0201E+0(7,+)}&\scriptsize{1.0270E+0(3,$\approx$)}&\scriptsize{1.0238E+0(6,+)}&\scriptsize{1.0241E+0(5,+)}&\scriptsize{1.0246E+0(4,$\approx$)}&\scriptsize{1.0274E+0(2,$\approx$)}\\\cline{3-9} 
&5&\scriptsize{\textbf{4.3619E-1(1)}}&\scriptsize{3.8177E-1(7,+)}&\scriptsize{4.3122E-1(2,+)}&\scriptsize{4.2359E-1(4,+)}&\scriptsize{4.0137E-1(6,+)}&\scriptsize{4.1566E-1(5,+)}&\scriptsize{4.2659E-1(3,+)}\\\cline{3-9} 
&8&\scriptsize{\textbf{5.2284E-2(1)}}&\scriptsize{2.8303E-2(7,+)}&\scriptsize{5.0318E-2(3,+)}&\scriptsize{5.1355E-2(2,+)}&\scriptsize{3.3162E-2(6,+)}&\scriptsize{4.9518E-2(4,+)}&\scriptsize{4.6619E-2(5,+)}\\\cline{3-9} 
&10&\scriptsize{\textbf{9.5250E-3(1)}}&\scriptsize{4.6711E-3(6,+)}&\scriptsize{9.1669E-3(2,+)}&\scriptsize{8.9611E-3(3,+)}&\scriptsize{4.2549E-3(7,+)}&\scriptsize{8.0157E-3(5,+)}&\scriptsize{8.2567E-3(4,+)}\\\hline 
\multirow{4}{*}{\makecell{Minus\\WFG6}}&3&\scriptsize{\textbf{1.0230E+0(1)}}&\scriptsize{1.0136E+0(7,+)}&\scriptsize{1.0227E+0(2,$\approx$)}&\scriptsize{1.0182E+0(6,+)}&\scriptsize{1.0214E+0(4,$\approx$)}&\scriptsize{1.0224E+0(3,$\approx$)}&\scriptsize{1.0209E+0(5,$\approx$)}\\\cline{3-9} 
&5&\scriptsize{\textbf{4.3608E-1(1)}}&\scriptsize{3.7684E-1(7,+)}&\scriptsize{4.2997E-1(2,+)}&\scriptsize{4.2475E-1(3,+)}&\scriptsize{3.9932E-1(6,+)}&\scriptsize{4.1790E-1(5,+)}&\scriptsize{4.2327E-1(4,+)}\\\cline{3-9} 
&8&\scriptsize{\textbf{5.3415E-2(1)}}&\scriptsize{2.7405E-2(7,+)}&\scriptsize{5.2042E-2(2,+)}&\scriptsize{5.2011E-2(3,+)}&\scriptsize{3.1316E-2(6,+)}&\scriptsize{4.9226E-2(5,+)}&\scriptsize{4.9490E-2(4,+)}\\\cline{3-9} 
&10&\scriptsize{\textbf{9.7459E-3(1)}}&\scriptsize{4.6892E-3(6,+)}&\scriptsize{9.4330E-3(2,+)}&\scriptsize{9.0637E-3(3,+)}&\scriptsize{3.8751E-3(7,+)}&\scriptsize{7.7272E-3(5,+)}&\scriptsize{8.9348E-3(4,+)}\\\hline 
\multirow{4}{*}{\makecell{Minus\\WFG7}}&3&\scriptsize{\textbf{1.0275E+0(1)}}&\scriptsize{1.0174E+0(7,+)}&\scriptsize{1.0247E+0(2,+)}&\scriptsize{1.0214E+0(5,+)}&\scriptsize{1.0214E+0(6,+)}&\scriptsize{1.0245E+0(4,+)}&\scriptsize{1.0245E+0(3,+)}\\\cline{3-9} 
&5&\scriptsize{\textbf{4.3709E-1(1)}}&\scriptsize{3.7840E-1(7,+)}&\scriptsize{4.3273E-1(2,+)}&\scriptsize{4.2472E-1(4,+)}&\scriptsize{3.9936E-1(6,+)}&\scriptsize{4.1509E-1(5,+)}&\scriptsize{4.2838E-1(3,+)}\\\cline{3-9} 
&8&\scriptsize{\textbf{5.3749E-2(1)}}&\scriptsize{1.9493E-2(7,+)}&\scriptsize{5.2113E-2(3,+)}&\scriptsize{5.2313E-2(2,+)}&\scriptsize{2.6106E-2(6,+)}&\scriptsize{4.8899E-2(4,+)}&\scriptsize{4.3529E-2(5,+)}\\\cline{3-9} 
&10&\scriptsize{\textbf{9.7260E-3(1)}}&\scriptsize{4.2562E-3(6,+)}&\scriptsize{9.4733E-3(2,+)}&\scriptsize{9.0393E-3(3,+)}&\scriptsize{2.9290E-3(7,+)}&\scriptsize{7.0638E-3(4,+)}&\scriptsize{6.7967E-3(5,+)}\\\hline 
\multirow{4}{*}{\makecell{Minus\\WFG8}}&3&\scriptsize{\textbf{1.0226E+0(1)}}&\scriptsize{1.0131E+0(7,+)}&\scriptsize{1.0217E+0(2,$\approx$)}&\scriptsize{1.0175E+0(6,$\approx$)}&\scriptsize{1.0189E+0(5,$\approx$)}&\scriptsize{1.0202E+0(3,$\approx$)}&\scriptsize{1.0202E+0(4,$\approx$)}\\\cline{3-9} 
&5&\scriptsize{\textbf{4.3732E-1(1)}}&\scriptsize{3.7507E-1(7,+)}&\scriptsize{4.3315E-1(2,+)}&\scriptsize{4.2417E-1(4,+)}&\scriptsize{3.9814E-1(6,+)}&\scriptsize{4.1572E-1(5,+)}&\scriptsize{4.2740E-1(3,+)}\\\cline{3-9} 
&8&\scriptsize{\textbf{5.4197E-2(1)}}&\scriptsize{2.6635E-2(7,+)}&\scriptsize{5.2890E-2(2,+)}&\scriptsize{5.2702E-2(3,+)}&\scriptsize{3.1459E-2(6,+)}&\scriptsize{4.9789E-2(5,+)}&\scriptsize{5.1372E-2(4,+)}\\\cline{3-9} 
&10&\scriptsize{\textbf{9.9498E-3(1)}}&\scriptsize{4.6947E-3(6,+)}&\scriptsize{9.6248E-3(2,+)}&\scriptsize{9.2396E-3(4,+)}&\scriptsize{3.9489E-3(7,+)}&\scriptsize{7.7568E-3(5,+)}&\scriptsize{9.4043E-3(3,+)}\\\hline 
\multirow{4}{*}{\makecell{Minus\\WFG9}}&3&\scriptsize{\textbf{1.0260E+0(1)}}&\scriptsize{1.0166E+0(7,+)}&\scriptsize{1.0237E+0(3,$\approx$)}&\scriptsize{1.0232E+0(4,+)}&\scriptsize{1.0222E+0(6,+)}&\scriptsize{1.0231E+0(5,+)}&\scriptsize{1.0250E+0(2,$\approx$)}\\\cline{3-9} 
&5&\scriptsize{\textbf{4.3708E-1(1)}}&\scriptsize{3.7353E-1(7,+)}&\scriptsize{4.3241E-1(2,+)}&\scriptsize{4.2603E-1(4,+)}&\scriptsize{3.9832E-1(6,+)}&\scriptsize{4.1289E-1(5,+)}&\scriptsize{4.3028E-1(3,+)}\\\cline{3-9} 
&8&\scriptsize{\textbf{5.1904E-2(1)}}&\scriptsize{2.7147E-2(7,+)}&\scriptsize{5.0250E-2(3,+)}&\scriptsize{5.1418E-2(2,+)}&\scriptsize{3.2505E-2(6,+)}&\scriptsize{4.9995E-2(4,+)}&\scriptsize{4.5941E-2(5,+)}\\\cline{3-9} 
&10&\scriptsize{\textbf{9.5579E-3(1)}}&\scriptsize{3.9006E-3(7,+)}&\scriptsize{9.1611E-3(2,+)}&\scriptsize{8.9784E-3(3,+)}&\scriptsize{4.2195E-3(6,+)}&\scriptsize{8.0962E-3(4,+)}&\scriptsize{7.9394E-3(5,+)}\\\hline 
\multicolumn{2}{|c|}{Average Rank}&$\mathbf{1.42}$&$6.75$&$2.54$&$3.62$&$5.63$&$4.33$&$3.71$\\ \hline 
\multicolumn{3}{|c|}{$+$/$-$/$\approx$}&51/0/1&38/2/12&42/4/6&47/1/4&41/4/7&42/2/8\\ \hline 
\end{tabular}
 \label{tab:Minus}\end{table*}

\section{Further Investigations}
\subsection{The Effect of the Training Solution Sets}
In Section \ref{section:lta},  we claimed that a single training solution set is not enough for LtA to train a good direction vector set. In this subsection, we perform a simple experiment to verify our claim. We follow the experimental settings in Section \ref{learninglta}. However, instead of using 100 solution sets $\{S_1,S_2,...,S_{100}\}$ for training, we use only a single solution set $S_1$. The LtA method using a single training solution set is denoted as LtA-S. The 3-objective case is considered for illustration.

\begin{figure}[!htbp]
\centering                                           
\subfigure[Learning curve]{ 
\includegraphics[scale=0.21]{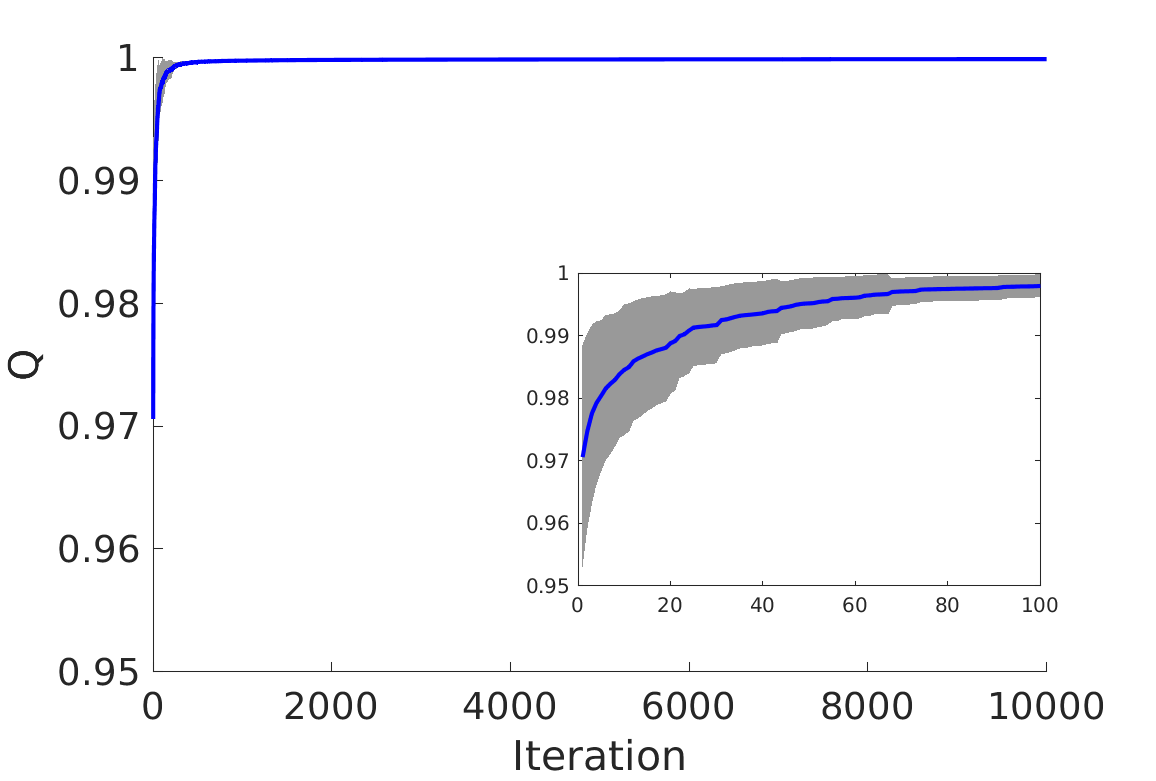}   }            
\subfigure[Learned direction vector set]{ 
\includegraphics[scale=0.3]{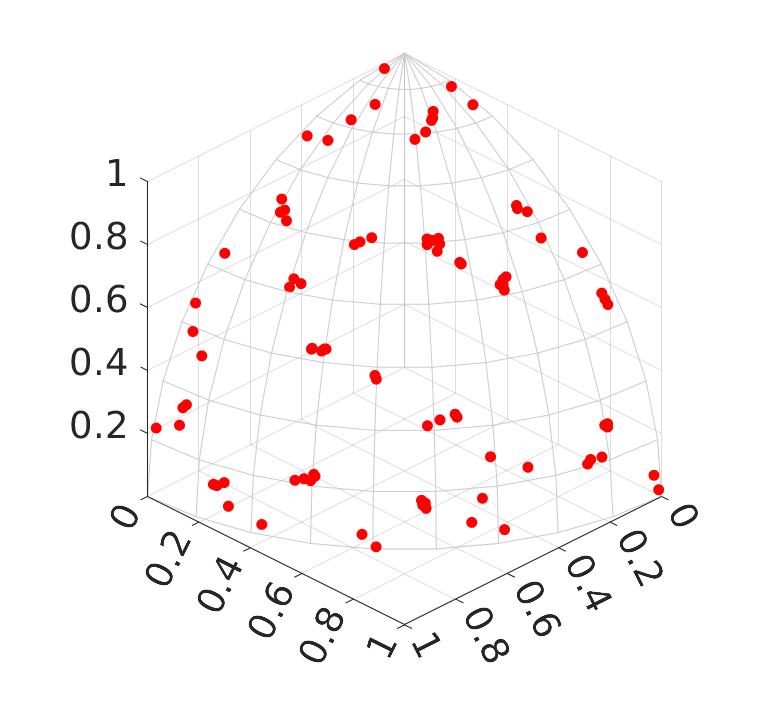} }
\caption{The learning curve of LtA-S over 20 independent runs and the learned direction vector set in a single run of LtA-S in 3-objective case.} 
\label{learningoneset}                                                        
\end{figure}

Fig. \ref{learningoneset} (a) shows the learning curve of LtA-S over 20 independent runs and Fig. \ref{learningoneset} (b) shows the learned direction vector set in a single run of LtA-S. We can see that a good $Q$ value is achieved by LtA-S at the end of the learning process. The learned direction vector set is highly non-uniform. 
\begin{figure}[!htbp]
\centering                                           
\includegraphics[scale=0.17]{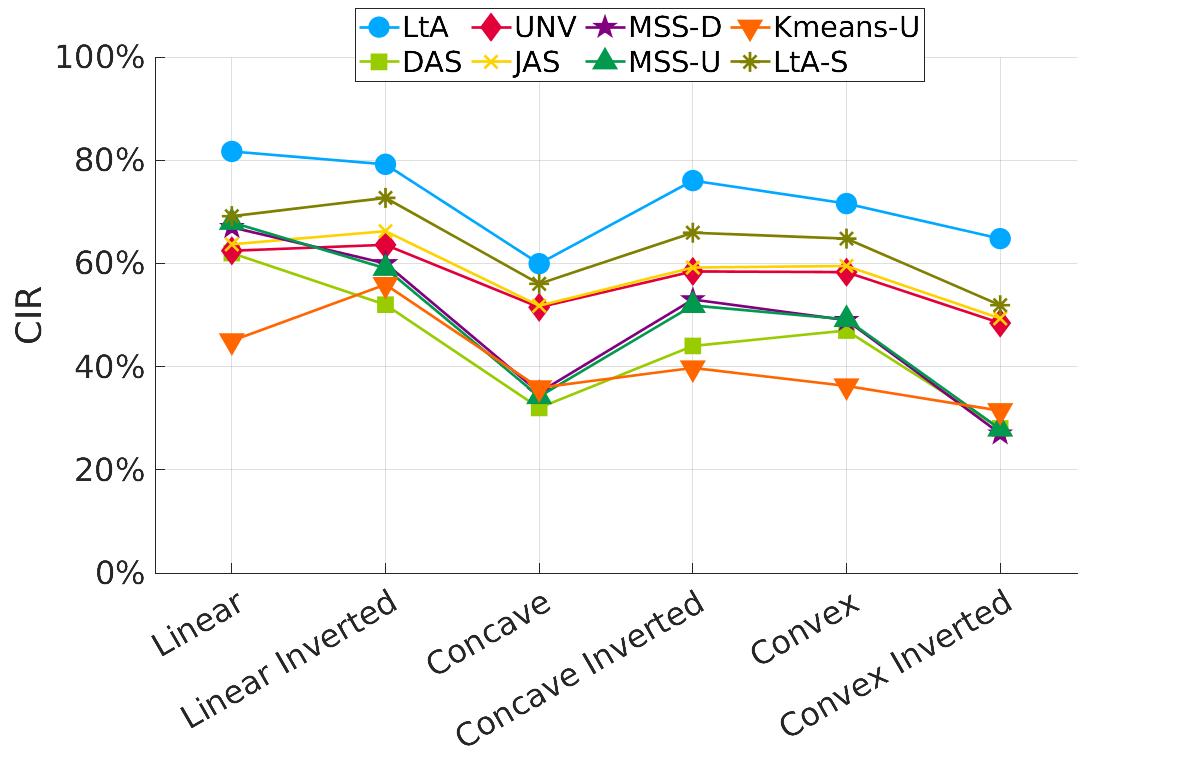}             
\caption{Correct identification rate (CIR) of the $R_2^{\text{HVC}}$ indicator with different direction vector sets on different solution sets in 3-objective case. All the results are based on 20 runs.} 
\label{further1}                                                        
\end{figure}

We perform experiment 1 (i.e., CIR) to evaluate the performance of the learned direction vector sets by LtA-S. Fig. \ref{further1} shows the comparison results among different methods. We can observe that compared with LtA, the performance of LtA-S is degraded. This shows that a single training solution set is not as good as a set of diverse training solution sets. However, interestingly, we can observe that LtA-S outperforms all the other methods. That is, even a single training solution set is used in LtA, better direction vector sets are obtained than the other methods. This reflects the usefulness of LtA for auto direction vector set generation.

\subsection{The Non-uniformity of the Learned Direction Vector Sets}
We have shown that the uniform direction vector sets generated by DAS, MSS-D, MSS-U, and Kmeans-U perform worse than the non-uniform direction vector sets generated by UNV, JAS, and LtA (except that Kmeans-U performs better than UNV and JAS in Experiment 2). It seems that a uniform direction vector set is not a good choice for the $R_2^{\text{HVC}}$ indicator.  In order to further verify this counterintuitive conclusion, we perform additional experiments to further examine whether the uniformity improvement of the learned direction vector set by LtA in Fig. \ref{learnedsets} (a) can improve its approximation quality. More specifically, we performed the following two experiments. (1) The learned direction vector set in Fig. \ref{learnedsets} (a) is used as the initial direction vector set and LtA is run for another 10,000 iterations. (2) We modify the learned direction vector set in Fig. \ref{learnedsets} (a) to make it more uniform. 

\begin{figure}[!htbp]
\centering                                           
\subfigure[Learning curve]{ 
\includegraphics[scale=0.21]{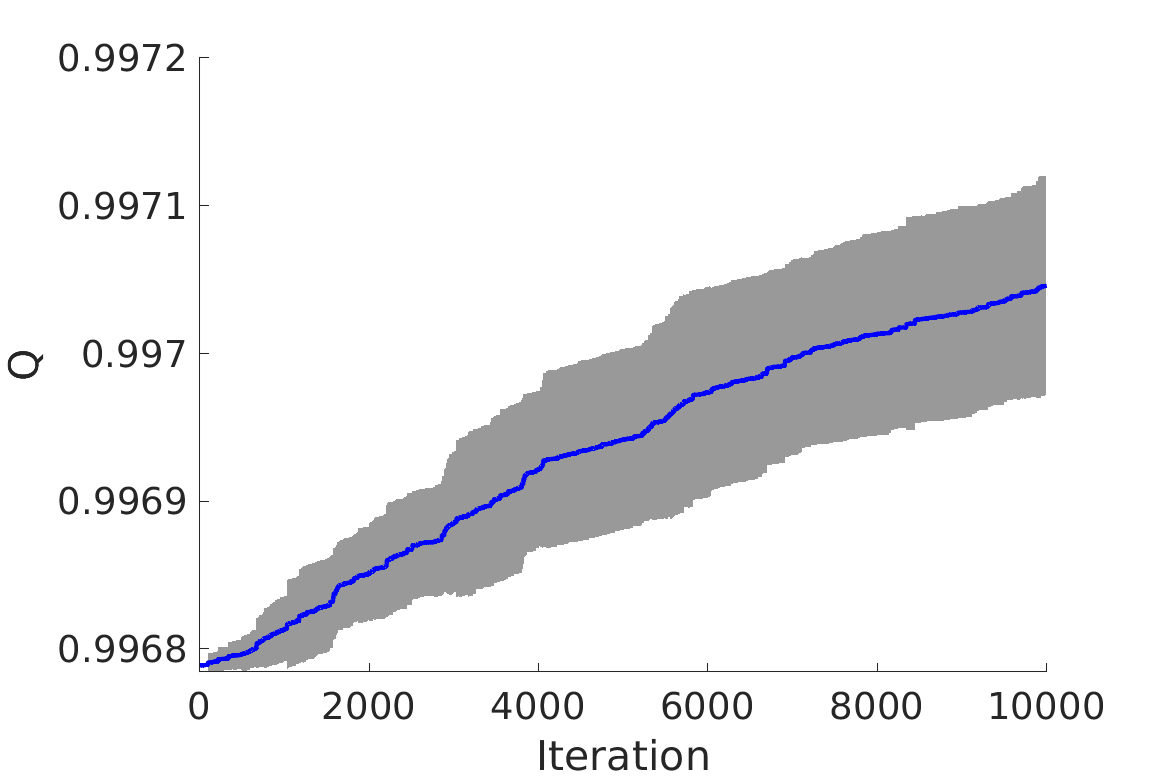}   }            
\subfigure[Learned direction vector set]{ 
\includegraphics[scale=0.3]{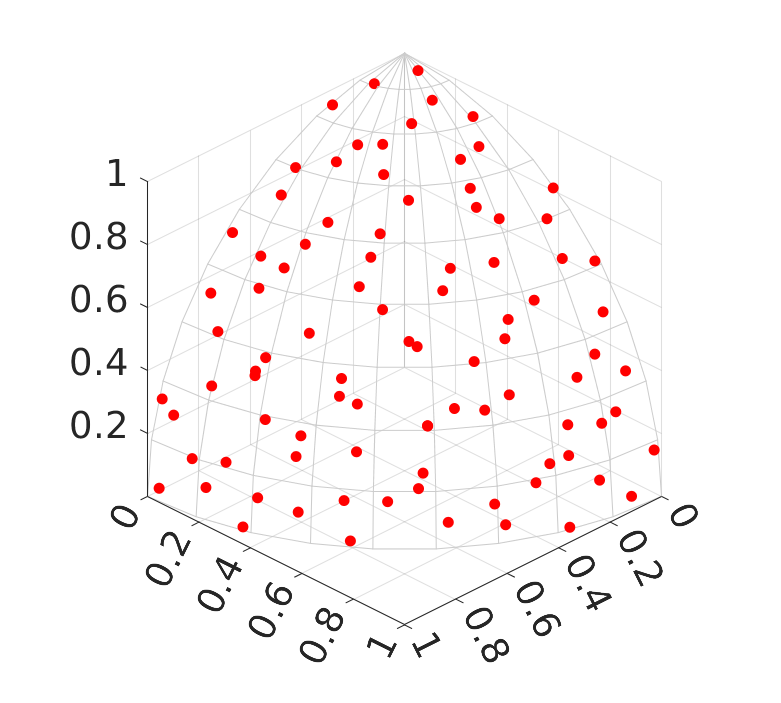} }
\caption{The learning curve of LtA over 20 runs when Fig. \ref{learnedsets} (a) is used as the initial direction vector set, and the learned direction vector set in a single run in 3-objective case.} 
\label{learningmore}                                                        
\end{figure}

Fig. \ref{learningmore} (a) shows the learning curve of LtA over 20 runs when Fig. \ref{learnedsets} (a) is used as the initial direction vector set, and Fig. \ref{learningmore} (b) shows the learned direction vector set in a single run. We can see that the improvement of the $Q$ value is very minor (less than 0.0003) and the learned direction vector set is still non-uniform after another 10,000 iterations. A uniform direction vector set cannot be obtained. 

\begin{figure}[!htbp]
\centering                                           
\subfigure[Original direction vector set]{ 
\includegraphics[scale=0.3]{AUTO} }          
\subfigure[Modified direction vector set]{ 
\includegraphics[scale=0.31]{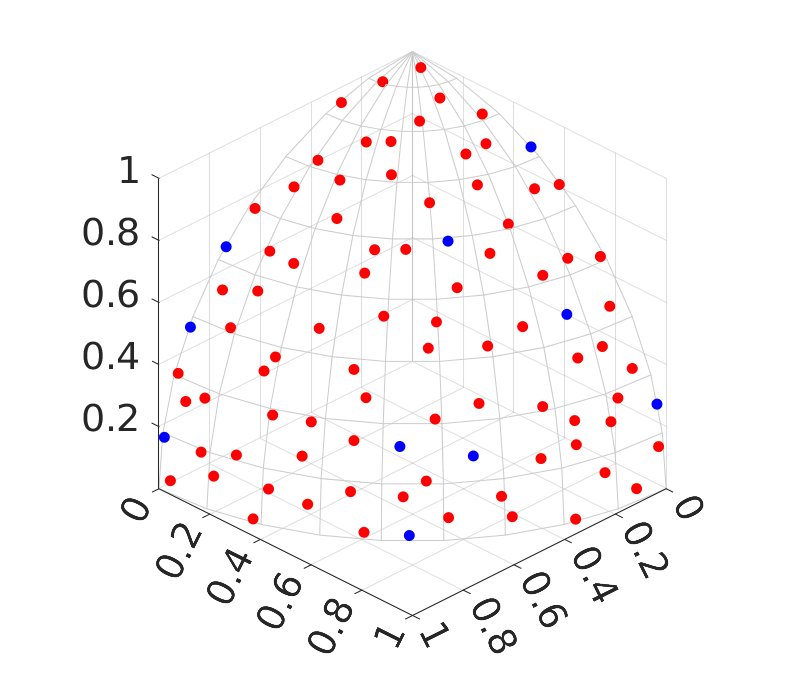} }
\caption{The original and modified direction vector sets based on Fig. \ref{learnedsets} (a). 10 out of 91 direction vectors are modified. The modified direction vectors are highlighted in blue in (b).} 
\label{manual}                                                        
\end{figure}
\begin{figure}[!htbp]
\centering                                           
\includegraphics[scale=0.17]{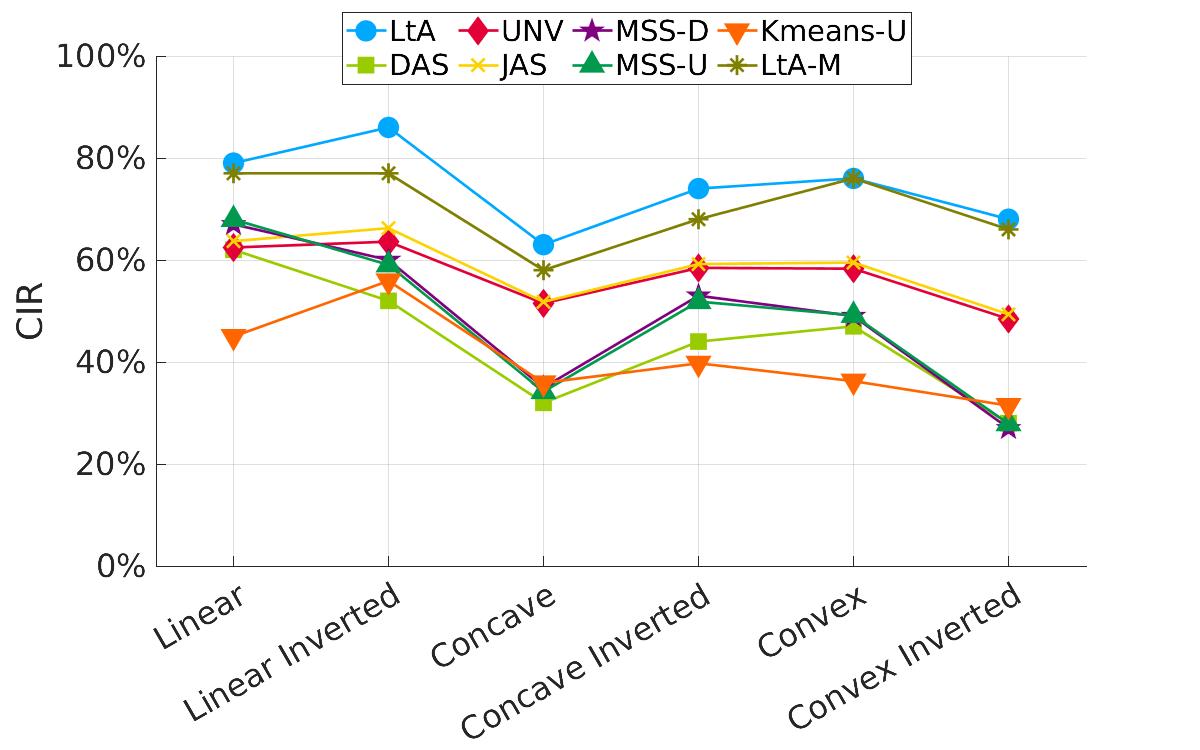}             
\caption{Correct identification rate (CIR) of the $R_2^{\text{HVC}}$ indicator with different direction vector sets on different solution sets in 3-objective case. The results of LtA and LtA-M are based on the two direction vector sets in Fig. \ref{manual}.} 
\label{manual1}                                                        
\end{figure}

Fig. \ref{manual} (a) shows the original direction vector set (the same as in Fig. \ref{learnedsets} (a)). Fig. \ref{manual} (b) shows the modified direction vector set where the modified direction vectors are highlighted in blue. We can see that the modified direction vector set is more uniform than before. We perform experiment 1 (i.e., CIR) to evaluate the performance of the modified direction vector set (denoted as LtA-M). Fig. \ref{manual1} shows the comparison results among different methods. Note that the results for LtA and LtA-M in Fig. \ref{manual1} are based on the two direction vector sets in Fig. \ref{manual} (whereas the results for LtA and LtA-S in Fig. \ref{further1} are based on 20 runs). We can observe that the performance of LtA-M is degraded. That is, although the uniformity of the direction vector set is improved, its performance becomes worse. However, we can see that LtA-M outperforms all the other methods. This shows that the slightly modified direction vector set from the learned one still has a good performance. 

These results support our conclusion: the learned direction vector set by the LtA method is non-uniform and has better approximation quality than uniformly generated direction vector sets.


\subsection{The Runtime of LtA}
\begin{figure}[]
\centering                                           
\subfigure[Part I]{ 
\includegraphics[scale=0.15]{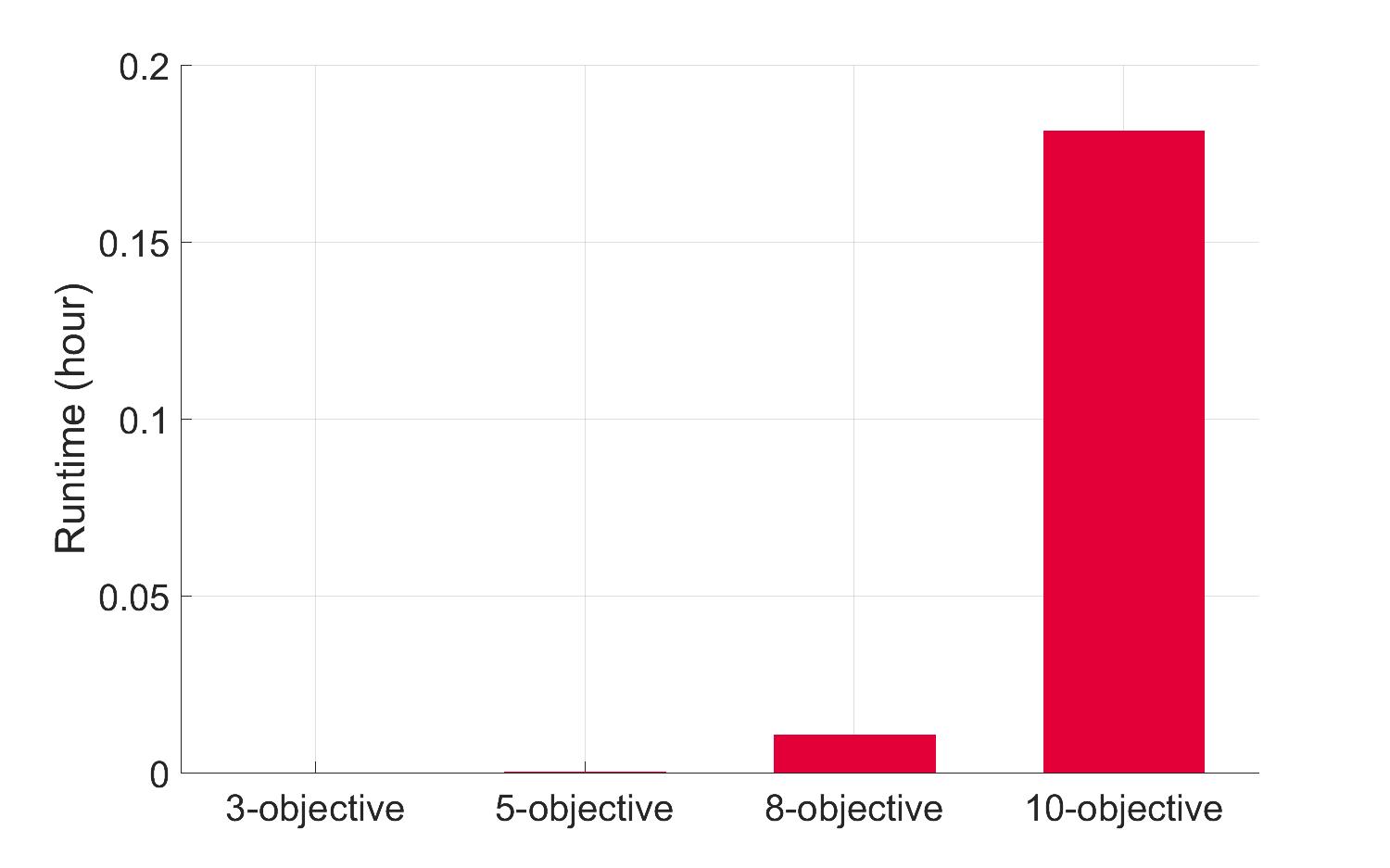} }          
\subfigure[Part II]{ 
\includegraphics[scale=0.15]{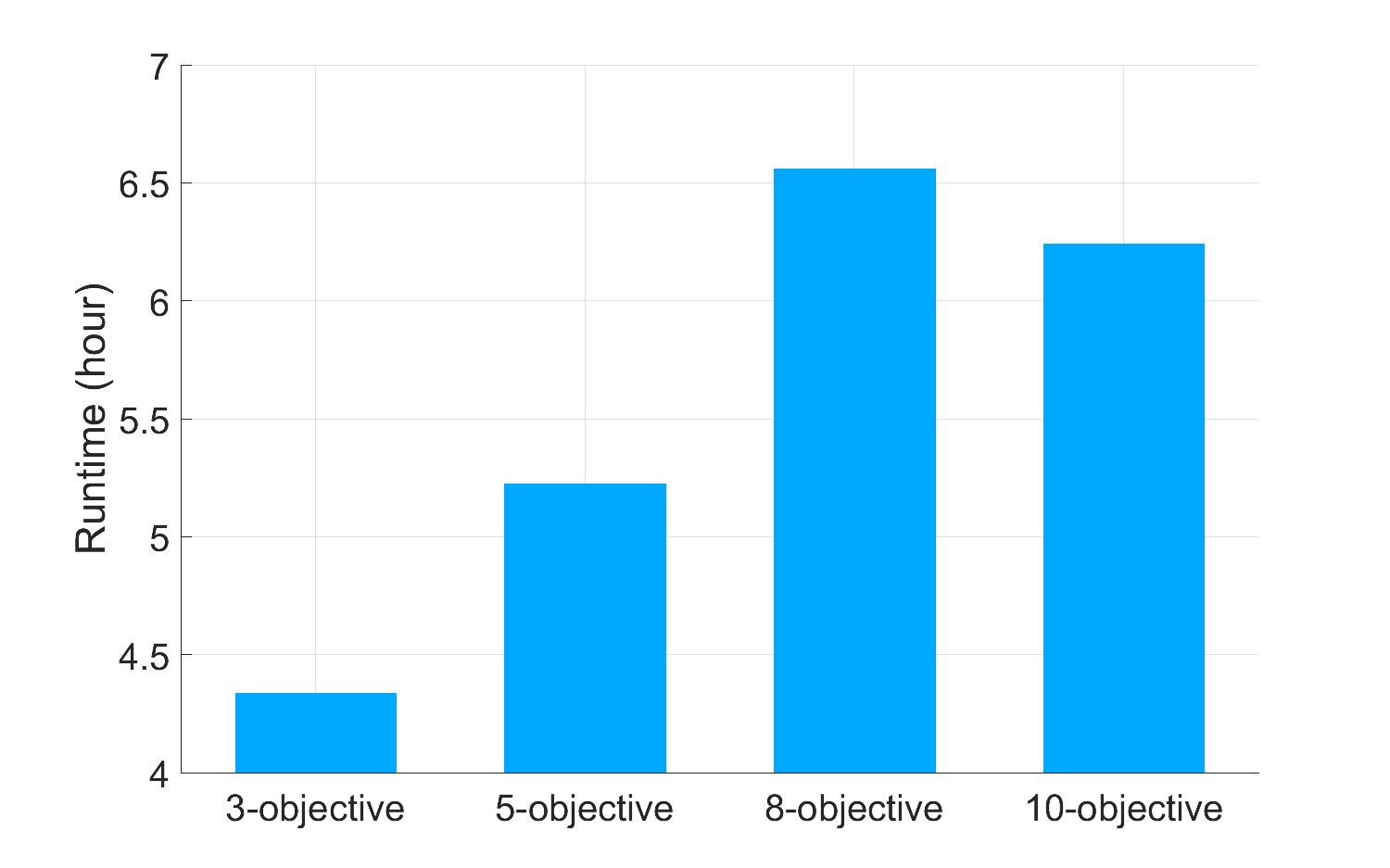} }\caption{The runtime of a single run of LtA in each objective case. (a) Part I: the calculation of the hypervolume contribution for the training solution sets (Line 1 in Algorithm 1). (b) Part II: the loop to iteratively learn the direction vector set (Lines 4-14 in Algorithm 1). The size of the direction vector set is 91, 105, 120 and 110 in 3, 5, 8 and 10-objective cases, respectively.} 
\label{runtime}                                                        
\end{figure}

We divide the runtime of LtA into two parts. Part I is the calculation of the hypervolume contribution for the training solution sets (Line 1 in Algorithm 1). Part II is the loop to iteratively learn the direction vector set (Lines 4-14 in Algorithm 1). We investigate the runtime of LtA by following the experimental settings in Section \ref{learninglta}. For the hypervolume contribution calculation, we use the WFG algorithm \cite{While2012A}. Our hardware platform is a virtual machine equipped with Intel Core i7-8700K CPU @3.70GHz. Fig. \ref{runtime} shows the runtime of a single run of LtA in each objective case. 

Fig. \ref{runtime} (a) shows the runtime of Part I (i.e., hypervolume contribution calculation). We can see that the runtime in 3 and 5-objective cases can be neglected, and is very minor in 8-objective case. It only takes relatively long runtime in 10-objective case. This is because the runtime of the hypervolume contribution calculation increases exponentially as the number of objectives increases. However, we do not need to perform Part I for each run if the same training solution sets are used among multiple runs. That is, we can save the hypervolume contribution information for further reuse.

Fig. \ref{runtime} (b) shows the runtime of Part II (i.e., learning process of LtA). We can see that it needs a little bit long runtime (more than 4 hours) to learn a direction vector set. As analyzed in Section \ref{timecomplexity}, the time complexity of LtA is proportional to the training solution set size, the direction vector set size, and the maximum iteration number. We can control the runtime of LtA by modifying these parameters. 

Although the learning process of LtA needs a little bit long runtime, the learned direction vector sets can be saved and are ready to use at any time. That is, once we obtain a good direction vector set by LtA, we can save it and use it directly in the future without spending a lot of time to regenerate it.

                     
\section{Conclusions}
The direction vector set plays an important role in the $R_2^{\text{HVC}}$ indicator for hypervolume contribution approximation. We investigated how to generate a good direction vector set for the $R_2^{\text{HVC}}$ indicator, and proposed a learning method of direction vectors for improving the approximation quality of $R_2^{\text{HVC}}$. The proposed method, LtA, is able to automatically learn a direction vector set based on given training solution sets. We verified the effectiveness of LtA based on three different experiments. All the results showed that LtA outperforms the other six direction vector set generation methods, which suggests the usefulness of LtA for auto direction vector set generation. 

In the future, we will consider to further improve the LtA method. One direction is to improve the initialization of the direction vector set. Another direction is to improve the generation of a new direction vector in each iteration. We will also consider to theoretically investigate the reason behind the non-uniformity of the learned direction vector sets by LtA.

All the data sets and results involved in our experiments are available at \url{https://github.com/HisaoLabSUSTC/LtA}. 


\bibliographystyle{IEEEtran}
\bibliography{sample-bibliography}

\begin{thebibliography}{10}
\providecommand{\url}[1]{#1}
\csname url@samestyle\endcsname
\providecommand{\newblock}{\relax}
\providecommand{\bibinfo}[2]{#2}
\providecommand{\BIBentrySTDinterwordspacing}{\spaceskip=0pt\relax}
\providecommand{\BIBentryALTinterwordstretchfactor}{4}
\providecommand{\BIBentryALTinterwordspacing}{\spaceskip=\fontdimen2\font plus
\BIBentryALTinterwordstretchfactor\fontdimen3\font minus
  \fontdimen4\font\relax}
\providecommand{\BIBforeignlanguage}[2]{{%
\expandafter\ifx\csname l@#1\endcsname\relax
\typeout{** WARNING: IEEEtran.bst: No hyphenation pattern has been}%
\typeout{** loaded for the language `#1'. Using the pattern for}%
\typeout{** the default language instead.}%
\else
\language=\csname l@#1\endcsname
\fi
#2}}
\providecommand{\BIBdecl}{\relax}
\BIBdecl

\bibitem{van1999multiobjective}
D.~A. Van~Veldhuizen, ``Multiobjective evolutionary algorithms:
  {C}lassifications, analyses, and new innovations,'' Ph.D. dissertation,
  Graduate School of Engineering, Air Force Institute of Technology, 1999.

\bibitem{coello2004study}
C.~A.~C. Coello and M.~R. Sierra, ``A study of the parallelization of a
  coevolutionary multi-objective evolutionary algorithm,'' in \emph{Mexican
  International Conference on Artificial Intelligence}.\hskip 1em plus 0.5em
  minus 0.4em\relax Springer, 2004, pp. 688--697.

\bibitem{zitzler2003performance}
E.~Zitzler, L.~Thiele, M.~Laumanns, C.~M. Fonseca, and V.~G. Da~Fonseca,
  ``Performance assessment of multiobjective optimizers: An analysis and
  review,'' \emph{IEEE Transactions on Evolutionary Computation}, vol.~7,
  no.~2, pp. 117--132, 2003.

\bibitem{shang2020survey}
K.~Shang, H.~Ishibuchi, L.~He, and L.~M. Pang, ``A survey on the hypervolume
  indicator in evolutionary multiobjective optimization,'' \emph{IEEE
  Transactions on Evolutionary Computation}, vol.~25, no.~1, pp. 1--20, 2021.

\bibitem{hansen1998evaluating}
M.~P. Hansen and A.~Jaszkiewicz, \emph{Evaluating the quality of approximations
  to the non-dominated set}.\hskip 1em plus 0.5em minus 0.4em\relax IMM,
  Department of Mathematical Modelling, Technical University of Denmark, 1998.

\bibitem{Emmerich2005An}
M.~Emmerich, N.~Beume, and B.~Naujoks, ``An {EMO} algorithm using the
  hypervolume measure as selection criterion,'' in \emph{2005 International
  Conference on Evolutionary Multi-Criterion Optimization}, 2005, pp. 62--76.

\bibitem{beume2007sms}
N.~Beume, B.~Naujoks, and M.~Emmerich, ``{SMS-EMOA}: Multiobjective selection
  based on dominated hypervolume,'' \emph{European Journal of Operational
  Research}, vol. 181, no.~3, pp. 1653--1669, 2007.

\bibitem{jiang2014simple}
S.~Jiang, J.~Zhang, Y.-S. Ong, A.~N. Zhang, and P.~S. Tan, ``A simple and fast
  hypervolume indicator-based multiobjective evolutionary algorithm,''
  \emph{IEEE Transactions on Cybernetics}, vol.~45, no.~10, pp. 2202--2213,
  2015.

\bibitem{bader2011hype}
J.~Bader and E.~Zitzler, ``{HypE}: An algorithm for fast hypervolume-based
  many-objective optimization,'' \emph{Evolutionary Computation}, vol.~19,
  no.~1, pp. 45--76, 2011.

\bibitem{shang2020new}
K.~Shang and H.~Ishibuchi, ``A new hypervolume-based evolutionary algorithm for
  many-objective optimization,'' \emph{IEEE Transactions on Evolutionary
  Computation}, vol.~24, no.~5, pp. 839--852, 2020.

\bibitem{bringmann2012approximating}
K.~Bringmann and T.~Friedrich, ``Approximating the least hypervolume
  contributor: {NP-hard} in general, but fast in practice,'' \emph{Theoretical
  Computer Science}, vol. 425, pp. 104--116, 2012.

\bibitem{bader2010faster}
J.~Bader, K.~Deb, and E.~Zitzler, ``Faster hypervolume-based search using
  {M}onte {C}arlo sampling,'' in \emph{Multiple Criteria Decision Making for
  Sustainable Energy and Transportation Systems}.\hskip 1em plus 0.5em minus
  0.4em\relax Springer, 2010, pp. 313--326.

\bibitem{shang2018r2}
K.~Shang, H.~Ishibuchi, and X.~Ni, ``R2-based hypervolume contribution
  approximation,'' \emph{IEEE Transactions on Evolutionary Computation},
  vol.~24, no.~1, pp. 185--192, 2020.

\bibitem{deng2019approximating}
J.~Deng and Q.~Zhang, ``Approximating hypervolume and hypervolume contributions
  using polar coordinate,'' \emph{IEEE Transactions on Evolutionary
  Computation}, vol.~23, no.~5, pp. 913--918, 2019.

\bibitem{nan2020good}
Y.~Nan, K.~Shang, and H.~Ishibuchi, ``What is a good direction vector set for
  the {R2}-based hypervolume contribution approximation,'' in \emph{Proceedings
  of the 2020 Genetic and Evolutionary Computation Conference}, 2020, pp.
  524--532.

\bibitem{das1998normal}
I.~Das and J.~E. Dennis, ``Normal-boundary intersection: A new method for
  generating the {P}areto surface in nonlinear multicriteria optimization
  problems,'' \emph{SIAM Journal on Optimization}, vol.~8, no.~3, pp. 631--657,
  1998.

\bibitem{zhang2007moea}
Q.~Zhang and H.~Li, ``{MOEA/D}: A multiobjective evolutionary algorithm based
  on decomposition,'' \emph{IEEE Transactions on evolutionary computation},
  vol.~11, no.~6, pp. 712--731, 2007.

\bibitem{deb2013evolutionary}
K.~Deb and H.~Jain, ``An evolutionary many-objective optimization algorithm
  using reference-point-based nondominated sorting approach, {P}art {I}:
  {S}olving problems with box constraints,'' \emph{IEEE Transactions on
  Evolutionary Computation}, vol.~18, no.~4, pp. 577--601, 2013.

\bibitem{jaszkiewicz2002performance}
A.~Jaszkiewicz, ``On the performance of multiple-objective genetic local search
  on the 0/1 knapsack problem - a comparative experiment,'' \emph{IEEE
  Transactions on Evolutionary Computation}, vol.~6, no.~4, pp. 402--412, 2002.

\bibitem{deb2019generating}
K.~Deb, S.~Bandaru, and H.~Seada, ``Generating uniformly distributed points on
  a unit simplex for evolutionary many-objective optimization,'' in
  \emph{International Conference on Evolutionary Multi-Criterion
  Optimization}.\hskip 1em plus 0.5em minus 0.4em\relax Springer, 2019, pp.
  179--190.

\bibitem{ishibuchi2018reference}
H.~Ishibuchi, R.~Imada, Y.~Setoguchi, and Y.~Nojima, ``Reference point
  specification in inverted generational distance for triangular linear
  {P}areto front,'' \emph{IEEE Transactions on Evolutionary Computation},
  vol.~22, no.~6, pp. 961--975, 2018.

\bibitem{macqueen1967some}
J.~MacQueen \emph{et~al.}, ``Some methods for classification and analysis of
  multivariate observations,'' in \emph{Proceedings of the Fifth Berkeley
  Symposium on Mathematical Statistics and Probability}, vol.~1, no.~14.\hskip
  1em plus 0.5em minus 0.4em\relax Oakland, CA, USA, 1967, pp. 281--297.

\bibitem{benesty2009pearson}
J.~Benesty, J.~Chen, Y.~Huang, and I.~Cohen, ``Pearson correlation
  coefficient,'' in \emph{Noise Reduction in Speech Processing}.\hskip 1em plus
  0.5em minus 0.4em\relax Springer, 2009, pp. 1--4.

\bibitem{auger2010theoretically}
A.~Auger, J.~Bader, and D.~Brockhoff, ``Theoretically investigating optimal
  $\mu$-distributions for the hypervolume indicator: First results for three
  objectives,'' in \emph{International Conference on Parallel Problem Solving
  from Nature}.\hskip 1em plus 0.5em minus 0.4em\relax Springer, 2010, pp.
  586--596.

\bibitem{shang2021greedy}
K.~Shang, H.~Ishibuchi, and W.~Chen, ``Greedy approximated hypervolume subset
  selection for many-objective optimization,'' in \emph{Proceedings of the
  Genetic and Evolutionary Computation Conference}, 2021, pp. 448--456.

\bibitem{bradstreet2007incrementally}
L.~Bradstreet, L.~While, and L.~Barone, ``Incrementally maximising hypervolume
  for selection in multi-objective evolutionary algorithms,'' in \emph{2007
  IEEE Congress on Evolutionary Computation}.\hskip 1em plus 0.5em minus
  0.4em\relax IEEE, 2007, pp. 3203--3210.

\bibitem{guerreiro2016greedy}
A.~P. Guerreiro, C.~M. Fonseca, and L.~Paquete, ``Greedy hypervolume subset
  selection in low dimensions,'' \emph{Evolutionary Computation}, vol.~24,
  no.~3, pp. 521--544, 2016.

\bibitem{deb2005scalable}
K.~Deb, L.~Thiele, M.~Laumanns, and E.~Zitzler, ``Scalable test problems for
  evolutionary multiobjective optimization,'' in \emph{Evolutionary
  Multiobjective Optimization}.\hskip 1em plus 0.5em minus 0.4em\relax
  Springer, 2005, pp. 105--145.

\bibitem{huband2006review}
S.~Huband, P.~Hingston, L.~Barone, and L.~While, ``A review of multiobjective
  test problems and a scalable test problem toolkit,'' \emph{IEEE Transactions
  on Evolutionary Computation}, vol.~10, no.~5, pp. 477--506, 2006.

\bibitem{ishibuchi2017performance}
H.~Ishibuchi, Y.~Setoguchi, H.~Masuda, and Y.~Nojima, ``Performance of
  decomposition-based many-objective algorithms strongly depends on {P}areto
  front shapes,'' \emph{IEEE Transactions on Evolutionary Computation},
  vol.~21, no.~2, pp. 169--190, 2017.

\bibitem{While2012A}
L.~While, L.~Bradstreet, and L.~Barone, ``A fast way of calculating exact
  hypervolumes,'' \emph{IEEE Transactions on Evolutionary Computation},
  vol.~16, no.~1, pp. 86--95, 2012.

\end{thebibliography}

\end{document}